\documentclass{article}

\usepackage{mathrsfs} %pour les lettres majuscules rondes ==> fonction \mathscr 

\usepackage[]{graphicx}    % Graphiques,  dvips  pdftexdraft
\usepackage[gen]{eurosym}
\usepackage{verbatim}
\usepackage{amsthm}% Commentaires
\usepackage{amsmath}
\usepackage{dsfont}
\usepackage{amssymb}
\usepackage{latexsym}		% Symboles mathmatiques
\usepackage{tabularx}
\usepackage{setspace}
\usepackage{listings}

\usepackage{arxiv}

\usepackage[utf8]{inputenc} % allow utf-8 input
\usepackage[T1]{fontenc}    % use 8-bit T1 fonts
\usepackage{hyperref}       % hyperlinks
\usepackage{url}            % simple URL typesetting
\usepackage{booktabs}       % professional-quality tables
\usepackage{amsfonts}       % blackboard math symbols
\usepackage{nicefrac}       % compact symbols for 1/2, etc.
\usepackage{microtype}      % microtypography
\usepackage{natbib}

\usepackage{float}
\usepackage{multirow,lscape,longtable}

\newcommand{\cay}[1]{[\citeauthor{#1} \citeyear{#1}]}
\newcommand{\cayNP}[1]{\citeauthor{#1} (\citeyear{#1})}

\title{Interpretabilité des modèles : état des lieux des méthodes et application à l'assurance}

\author{
  Dimitri Delcaillau \\
  P\&C, Predictive Analytics\\
  Milliman\\
  Paris, France \\
  \texttt{dimitri.delcaillau@milliman.com} \\
   \And
 Antoine Ly \\
  Data Analytics Solutions\\
  SCOR \\
  Paris, France \\
  \texttt{aly@scor.com} \\
 \And
 Franck Vermet \\
  Laboratoire de Mathématiques, EURIA \\                          
  Université de Bretagne Occidentale\\
  Brest, France \\
  \texttt{franck.vermet@univ-brest.fr} \\ 
 \And
 Alizé Papp \\
  Data Analytics Solutions \\
  SCOR \\
  Charlotte, \'Etats-Unis \\
  \texttt{alize.papp@outlook.com} \\
}

\begin{document}
\maketitle

\begin{abstract}
Le Réglement Général sur la Protection des Données (RGPD) introduit depuis mai 2018 certaines obligations aux industries. En fixant un cadre légal, elle impose notamment une forte transparence sur l'usage des données à caractère personnel. Ainsi, un usager se doit d'être informé de l'exploitation de ses données et d'y consentir. Les données sont la matière première de nombreux modèles qui permettent aujourd'hui d'augmenter la qualité et la performances des services digitaux. La transparence sur l'usage de la donnée impose  également une bonne compréhension de son usage aux travers différents modèles. L'usage de modèles, aussi si performants soient-ils, doit s'accompagner d'une compréhension à tout niveau du processus de transformation de la donnée (en amont et en aval d'un modèle), permettant ainsi de définir les relations entre la donnée associée à un individu et le choix qu'un algorithme pourrait faire sur la base de l'analyse de ce dernier\footnote{Par exemple, la recommandation d'un produit, une offre promotionnelle ou encore un tarif assurantiel représentatif du risque}. Il doit être possible de s'assurer qu'un modèle n'effectue pas de discrimination et qu'il est également possible d'expliquer son résultat. L'élargissement du panel d'algorithmes prédictifs -rendu possible par l'évolution des capacités de calcul- amène les scientifiques à être vigilants sur l'utilisation des modèles et à considérer de nouveaux outils pour mieux comprendre les décisions déduites à partir de ces derniers. Récemment, la communauté s'est particulièrement activée sur le sujet de la transparence des modèles avec une intensification prononcée des publications ces trois dernières années. L'utilisation de plus en plus fréquente d'algorithmes plus complexes (\textit{deep learning}, Xgboost, etc.) présentant des performances séduisantes est sans doute l'une des causes de cet intérêt. Cet article présente ainsi un état des lieux des méthodes d'interprétation des modèles et leurs usages dans un contexte assurantiel.

\end{abstract}

% keywords can be removed

\section{L'interprétabilté des modèles : un enjeu majeur} 
\label{sec:0_interpretabilite}
\subsection{Définir l'interprétabilité}

\begin{figure}[H]
\begin{center}
\includegraphics[scale=0.39]{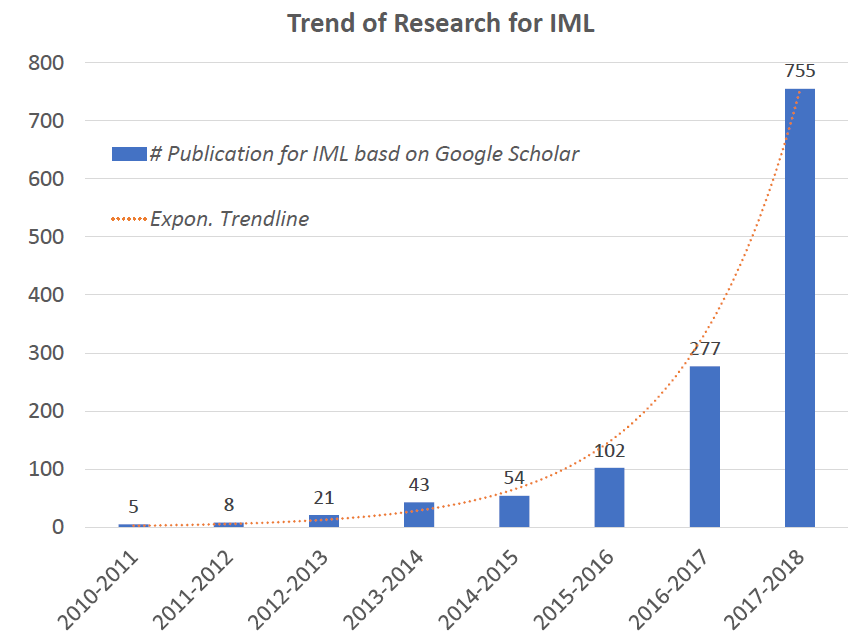}
\end{center}
\caption{Nombre d'articles publiés en lien avec l'interprétabilité des modèles de machine learning au cours des 15 dernières années  \cay{Peeking_Black_Box_XAI} }
    \label{fig:nb_papers_interpretability}
\end{figure}

Si l'on se réfère aux différentes publications de ces dernières années, l'interprétabilité est un nouvel enjeu dans l'utilisation des modèles et plus particulièrement ceux de \textit{machine learning}. \cayNP{Peeking_Black_Box_XAI} ont en effet montré l'intérêt croissant de la communauté scientifique et des régulateurs pour l'interprétation des modèles. Cependant, bien que le concept d'interprétabilité semble de plus en plus répandu, on note l'absence d'un consensus général tant sur la définition que sur la mesure de l'interprétabilité d'un modèle \cay{molnar2019}. Il existe effectivement de nombreuses méthodes (graphiques, mathématiques, etc.) qui peuvent être associées à la l'interprétation des algorithmes, ce qui entraîne parfois une certaine confusion autour de la notion. Par ailleurs, il peut décrire des degrés différents de compréhension selon la population visée : parlons-nous de la compréhension du modèle, de la capacité à contrôler les résultats de ce dernier, de sa transparence vis-à-vis d'utilisateurs novices ? Ou faisons-nous référence aux moyens mis en place pour analyser les résultats d'un algorithme aussi complexe soit-il ? \\ 

\cayNP{define_interpretability_ML} tentent de donner une définition précise à l'interprétabilité dans le cadre d'un modèle de machine learning. Ils fournissent notamment un cadre (appelé PDR) construit sur trois propriétés souhaitées pour l'évaluation et la construction d'une interprétation. Ce cadre est détaillé ci-aprés et permet de classer les différentes méthodes existantes et d'utiliser un vocabulaire commun entre les différents acteurs du domaine de l'apprentissage statistique.\\

\subsection{Définir les critères d'un modèle interprétable}

L'article \cay{define_interpretability_ML} suggère tout d'abord que l'interprétation fait référence à la notion d'extraction d'informations. \cayNP{miller} propose plus précisément de définir l'interprétabilité comme le degré à partir duquel un humain peut comprendre la cause d'une décision. Une définition alternative est également proposée par \cayNP{NIPS2016_6300} et reprise par \cayNP{molnar2019}: l'interprétabilité est définie comme \guillemotleft le degré à partir duquel un humain peut régulièrement prédire le résultat du modèle \guillemotright. Ainsi, une connaissance est dite pertinente \cay{define_interpretability_ML} si elle fournit une information pour un public particulier et un problème d'un domaine choisi. La notion d'interprétabilité peut donc s'évaluer selon plusieurs critères :

\begin{itemize}
    \item \textbf{La confiance} : ce critère revient régulièrement lorsque l'interprétabilité des modèles est abordée. Par exemple, \cay{lime} font référence à cet enjeu: lorsque l'on considère qu'un modèle fournit des résultats qu'un humain peut utiliser pour prendre des décisions, il apparaît clairement qu'il doit pouvoir s'appuyer sur le modèle en toute sérénité. 
    \item \textbf{La causalité} : bien que l'un des objectifs des algorithmes d'apprentissage statistique soit de mettre en avant des corrélations, l'algorithme doit permettre de mieux comprendre des phénomènes du monde réel et les interactions entre différents facteurs observés.
    \item \textbf{\guillemotleft La transférabilité\guillemotright} : définie comme la capacité d'un modèle à s'adapter à des situations légèrement différentes, elle est une des propriétés souhaitées dans la recherche d'interprétabilité. Elle transcrit en outre, la capacité de généralisation.
    \item \textbf{\guillemotleft L'informativité\guillemotright} : l'utilisation d'un algorithme doit pouvoir dépasser la simple optimisation mathématique. Un modèle doit pouvoir fournir une information précise sur sa prise de décision.
    \item \textbf{Une prise de décision juste et éthique} : ce critère rejoint les directives du RGPD. L'utilisateur d'un algorithme doit pouvoir s'assurer de l'absence de biais dans la prise de décision et un respect de l'éthique (ne pas commettre de discrimination par exemple).
\end{itemize}

Même s'il est difficile de prendre en compte objectivement l'ensemble de ces axes, l'interprétabilité peut être garantie par l'utilisation de deux grandes familles d'outils : ceux qui s'appuient sur le modèle lui-même et ceux qui s'appliquent \textit{a posteriori} via des analyses \textit{post-hoc}. Pour bien les choisir, le cadre PDR (\textit{Precision}, \textit{Description}, \textit{Relevance}) \cay{define_interpretability_ML} -ou PDP en français- suggère d'utiliser des méthodes qui permettent d'interpréter un modèle sous trois angles : la \guillemotleft précision prédictive\guillemotright, la \guillemotleft précision descriptive\guillemotright \ et la \guillemotleft pertinence\guillemotright.\\

 La précision dite prédictive (P) évalue l'aptitude predictive d'un modele en prêtant attentation non seulement à son efficacité en moyenne, comme il est de tradition en machine learning, mais aussi à sa distribution. Il peut en effet être problématique que l'erreur de prédiction soit bien plus élevée pour une classe spécifique, meme si le modele est trés prédictif en moyenne ; ou bien l'on peut vouloir s'assurer que le modèle est robuste (i.e. non sensible à l'échantillonnage) pour faire confiance aux relations qu'il met en avant.\\

La précision dite descriptive (D) est définie dans \cay{define_interpretability_ML} comme le \guillemotleft degré à partir duquel une méthode d'interprétation capture objectivement les relations apprises par les modèles de machine learning \guillemotright. En général, les modèles perçus comme interprétables comme les arbres de classification par exemple, ou les régression linéaire, possèdent des méthodes d'interprétation s'appuyant sur les modèles (graphiques, effet-marginaux, odd-ratio, etc.) avec une bonne précision descriptive. Lors du choix du modèle de \textit{machine learning} pour répondre à une problématique, un compromis est donc à réaliser entre précision prédictive et précision descriptive. Intrinsèquement, les modèles "simples" possèdent des méthodes d'interprétation à fort pouvoir descriptif alors que les modèles complexes (à forte paramétrisation comme le \textit{deep learning} ou à grande profondeur comme le Xgboost) ont une bonne précision prédictive mais nécessitent  des interprétations \textit{post-hoc} pour d'augmenter leur précision descriptive.\\

\begin{figure}[!h]
    \centering
    \includegraphics[width=55mm]{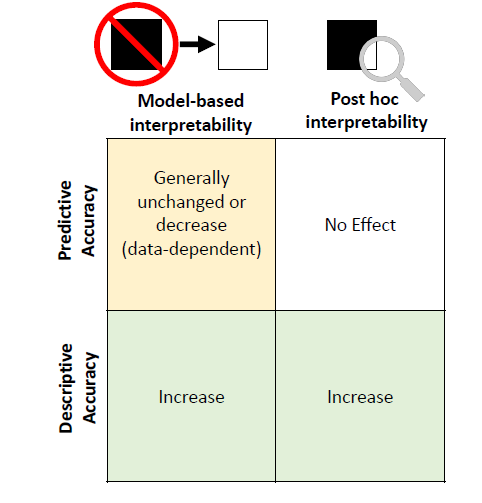}
    \caption{Impact des méthodes d'interprétabilité sur les précisions prédictive et descriptive dans le cadre du PDR  \cay{define_interpretability_ML}}
    \label{fig:PDR_acc}
\end{figure}

Enfin, la pertinence (R) de l'information apportée par l'interprétation du modèle est également cruciale. \cay{define_interpretability_ML} la définissent comme suit : \guillemotleft une interprétation est dite pertinente si elle fournit des informations pour un public particulier et un domaine choisi \guillemotright. L'interprétabilité est donc également dépendante du public concerné par le modèle : un décisionnaire, un médecin, un patient, un assuré, etc. Ce dernier critère permet alors parfois d'arbitrer entre la précision \guillemotleft prédictive \guillemotright et \guillemotleft Descriptive \guillemotright. L'interprétabilité se mesure donc par une approche méthodique et le choix d'outils adaptés afin d'assurer la bonne compréhension des résultats obtenus via un processus de modélisation. Comme introduit précedemment, l'interprétabilité peut donc s'étudier dans un premier temps sous deux niveaux : l'interprétabilité basée sur les modèles et l'interprétabilité \textit{post-hoc} (agnostique aux modèles).

\subsection{Les deux grands types d'interprétabilité}

\subsubsection{L'interprétabilité basée sur le modèle (IBM)}

L'interprétabilité basée sur le modèle (IBM), constitue le premier niveau d'interprétabilité. Elle intervient pendant l'élaboration du modèle et est liée au choix des familles d'algorithmes utilisées pour comprendre un phénomène et leur calibrage. Un modèle interprétable peut alors se définir par sa :

\begin{itemize}
    \item parcimonie : La parcimonie est étroitement liée au principe du rasoir d'Ockham, qui stipule que \guillemotleft les multiples ne doivent pas être utilisés sans nécessité \guillemotright. Dans le cas d'un modèle de \textit{machine learning}, imposer que le modèle soit parcimonieux revient à limiter le nombre de paramètres non nuls. En statistique comme en apprentissage automatique, il existe différentes méthodes de régularisation, applicables à de nombreux modèles\footnote{Le Xgboost introduit une méthode de régularisation tout comme le \textit{deeplearning} avec le \textit{drop-out}. Cependant même sous contrainte, ces modèles sont souvent peu parcimonieux.}. 
    \item simulabilité : \cayNP{define_interpretability_ML} définit un modèle comme simulable si un humain est capable de reproduire le processus de décision global de l'algorithme. Ainsi la simulabilité réfère à une transparence totale du modèle : un humain devrait être capable, à partir des entrées et des paramètres du modèle, de réaliser l'ensemble des calculs, en temps raisonnable, pour reconstruire la prédiction faite par le modèle. En ce sens, les arbres de décision sont généralement cités comme des algorithmes simulables, étant donné leur simplicité visuelle pour la prise de décision. De même, les règles de décision sont rangées dans cette catégorie.
    \item modularité : un modèle est modulaire si une portion significative du processus de prédiction peut être interprétée indépendamment. Ainsi, un modèle modulaire ne sera pas aussi simple à comprendre qu'un modèle parcimonieux ou simulable mais peut augmenter la précision descriptive en fournissant des relations apprises par l'algorithme. Un exemple classique de modèle considéré comme modulaire est la famille des GAM (modèles additifs généralisés) \cay{tibshirani1990generalized}, dont les GLM (régressions linéaires généralisées) sont une sous-famille. Dans ce type de modèles, la relation entre les variables est forcément additive et les coefficients trouvés permettent une interprétation relativement facile du modèle. Par opposition, les réseaux de neurones profonds sont eux considérés comme peu modulaires, étant donné le peu d'informations fournies par les coefficients de chaque couche.
    Dans une étude réalisée par Caruana et Al. (2015), il est prouvé que la probabilité de décès à cause de la pneumonie plus faible lorsque le patient est atteint d'asthme. 
    Cela vient du fait que les patients atteints d'asthme reçoivent un traitement plus agressif.  Si l'on suivait les préconisations données par l'algorithme, c'est-à-dire de rendre le traitement moins agressif pour les personnes atteintes d'asthme, le modèle deviendrait faux.
    Cet exemple montre l'intérêt de la modularité pour produire des interprétations pertinentes, de sorte à pouvoir détecter ensuite des biais dans la base d'apprentissage.
\end{itemize}

 Selon sa nature, un modèle possède des propriétés et outils d'analyse qui permettent une compréhension plus ou moins simple selon les points mentionnés précédemment. Le second niveau d'interprétation est moins sensible aux algorithmes utilisés lors du processus de modélisation.

\subsubsection{L'interprétabilité \textit{post-hoc}}

L'interprétabilité \textit{post-hoc}, à la différence de l'interprétabilité basée sur le modèle, correspond à l'analyse une fois que le modèle a été ajusté. Cette interprétation \textit{a posteriori} intervient dans le but de fournir des informations sur les relations éventuelles capturées par l'algorithme. C'est  sur ce type d'interprétation que la recherche a été particulièrement active ces dernières annnées. Elles s'avèrent particulièrement utiles pour analyser des modèles complexes mais à forte précision prédictive.\\

L'analyse \textit{post-hoc} vient augmenter la précision descriptive du modèle. Elle intervient plus particulièrement à deux niveaux : sur la compréhension du modèle au regard des données utilisées et sur l'analyse des prédictions fournies par l'algorithme. Elle est donc un supplément aux modèles utilisés. Ces méthodes ont connu ces dernières années des évolutions assez prononcées permettant de dépasser les limites des outils pré-existants d'analyse notamment ceux des arbres \cay{Breiman2001} ou des réseaux de neurones \cay{olden2004accurate}. \\

L'interprétation au niveau des données permet de s'intéresser aux relations générales apprises par le modèle, c'est-à-dire aux règles pertinentes d'une classe particulière de réponses ou d'une sous-population. De ces deux niveaux d'interprétation \textit{post-hoc} se dégagent des outils d'interprétation globaux et locaux. La section qui suit présente différentes méthodes d'interprétation \textit{post-hoc}.

\section{Les méthodes d'interprétation \textit{post-hoc}} 
\label{sec:1_methodesInterpretation}
\begin{figure}[H]
    \centering
    \includegraphics[scale=0.7]{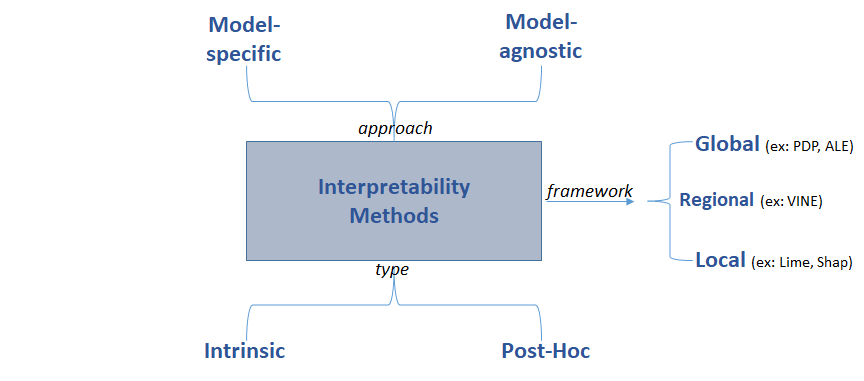}
    \caption{Différentes catégories d'interprétabilité des modèles}
    \label{scope_interpretability}
\end{figure}

Dans cette section nous nous intéressons aux méthodes \textit{post-hoc} agnostiques aux modèles. Il existe bien évidemment des méthodes propres à chaque algorithme renforçant l'interprétation de ces derniers (comme par exemple l'importance des variables des arbres \cay{Breiman2001}) mais ne sont pas l'objet de ce chapitre. Le schéma \ref{scope_interpretability} résume la répartition des méthodes d'interprétation selon différents niveaux d'un processus de modélisation. Nous distinguons en particulier différents cadres d'application de ces méthodes. Les deux grandes familles présentées dans la littérature sont les méthodes globales et locales. Ces dernières reposent sur la compréhension de la prédiction de la boîte-noire d'une observation en particulier alors que l'approche globale essaie de comprendre le modèle dans son intégralité. A mi-chemin entre ces deux familles se trouve le cadre régional, qui essaie d'expliquer le comportement du modèle pour un groupe d'individus similaires, par exemple à partir de clusters. Nous détaillons dans les sections suivantes les outils d'interprétation qui nous ont semblé les plus pertinents au regard des modèles complexes les plus couramment utilisés en assurance, à savoir les modèles par arbres (Random Forest ou Gradient Boosting). Ces méthodes sont par ailleurs applicables à tout algorithme (elles sont pour cela dites "agnostiques").

\subsection{Méthodes graphiques d'interprétation}

\subsubsection{Graphique de dépendance partielle (PDP)}
\label{subsec:PDP}

\paragraph{Présentation de la méthode}
L'analyse PDP (Partial Dependance Plot), introduite par \cay{pdp} est sans doute la méthode la plus ancienne d'interprétation des modèles au regard des publications de ces trois dernières années. Cette méthode graphique de dépendance partielle a pour objectif de montrer l'effet marginal d'une ou plusieurs variables explicatives sur la prédiction faite par un modèle. C'est une méthode d'interprétation globale.\\

Considérons une base d'apprentissage  $A_n=\{Z^{(i)}=(X^{(i)},Y^{(i)}) \in \mathds{R}^p$x$\mathds{R}, i=1,\dots, n\}$ constituée de $n$ vecteurs aléatoires indépendants et de même loi et  un modèle $\hat{f}$ entraîné sur des observations $(x^{(i)},y^{(i)})$ de la base $A_n$, avec $x^{(i)}=(x_j^{(i)})_{1 \leq j \leq p}$ pour $\ 1  \leq i \leq n$ . Notons $X_S$ l'ensemble des variables pour lesquelles on veut connaître l'effet sur la prédiction et $X_C$ les variables explicatives restantes. Par exemple $X_S=(X_1,X_2)$ et $X_C=(X_3,...,X_p)$. Ainsi, $X=(X_S,X_C)$, représente l'ensemble des variables explicatives utilisées par notre modèle.\\

On définit alors la fonction de dépendance partielle par la formule: \begin{equation}\hat{f}_{x_S}(x_S)=\mathds{E}_{X_c}[\hat{f}(x_S,X_C)] = \int\hat{f}(x_s,x_c)d\mathds{P}_{X_c}(x_c)\end{equation}

Notons que cette formule diffère de l'espérance conditionnelle de $X_S$. Afin de l'estimer, il suffit d'utiliser les $n$ observations et la méthode de Monte Carlo pour estimer l'espérance :

\begin{equation}\hat{f}_{x_S}(x_S) \simeq \frac{1}{n} \sum\limits_{i=1}^{n}{\hat{f}(x_S,x_C^{(i)})} \end{equation}

\begin{figure}[!h]
    \centering
    \includegraphics[width=120mm]{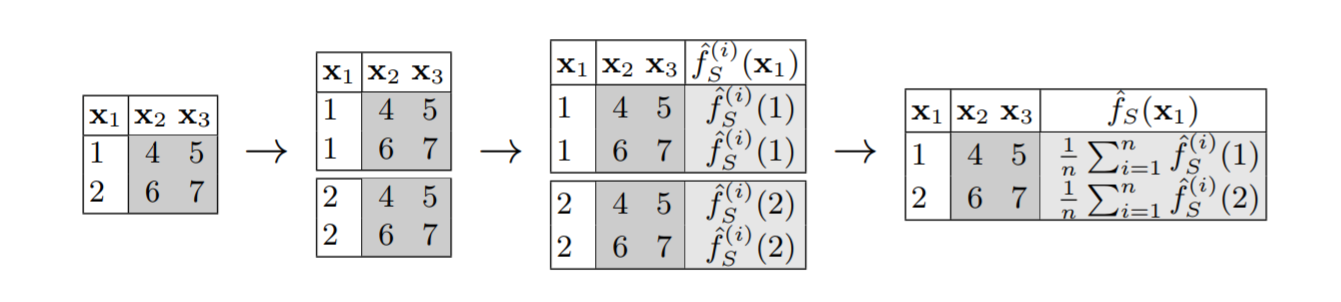}
    \caption{Calcul du graphique PDP sur un exemple simple}
    \label{fig:PD_calcul}
\end{figure}

L'algorithme de construction de la courbe est détaillé en annexe \ref{algo_pdp}. Il repose sur l'hypothèse forte de non corrélation entre les variables de l'ensemble $C$ et celles de $S$. Dans la pratique, ce cas est rarement vérifié ce qui mène à la considération d'associations de modalités non possibles en réalité (par exemple d'observer un individu de 2m avec un poids inférieur à 10kg si on considère des variables morphologiques). \\

\begin{figure}[!h]
\begin{tabular}{ccc}
\centering
\includegraphics[width=60mm]{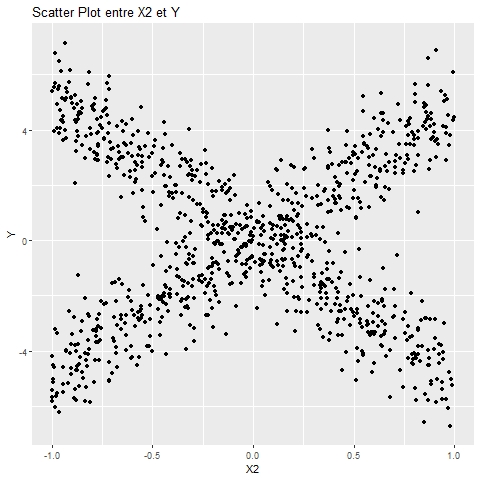} &

\includegraphics[width=80mm]{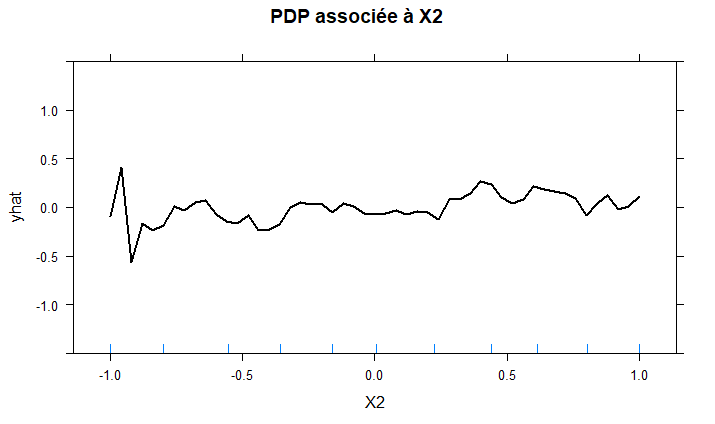} 
\end{tabular}
\caption{Scatter Plot de $X_2$ et $Y$ (à gauche) et graphique PDP de $X_2$ (à droite)}
    \label{fig:scatterPDP}
\end{figure}

\paragraph{Illustration}
Afin d'illustrer la méthode, considérons l'exemple qui introduit les variables ci-après.

$$Y=0.2X_1-5X_2+10 X_2 \mathds{1}_{\{ X_3 \geq 0\}}+ \varepsilon,$$ où $X_1,X_2,X_3 \overset{iid}{\sim}{U(-1,1)}, \varepsilon \sim N(0,1) $ avec $\varepsilon$ indépendant de $X_1,X_2,X_3$.
Supposons que nous observons un échantillon de taille $n=1000$.
Le nuage de points (\emph{scatter-plot}) associé à $X_2$ et $Y$ de cet échantillon est représenté à gauche de la figure \ref{fig:scatterPDP}. Le graphique de PDP de la variable $X_2$ associé au modèle de \textit{Random Forest} mis en place  afin de prédire $Y$ est représenté à droite de la figure \ref{fig:scatterPDP}.

\

Le graphique \ref{fig:scatterPDP} de PDP suggère qu'en moyenne la variable $X_2$ n'est pas significative dans la prédiction de $Y$ par le modèle de forêt aléatoire alors que le nuage de points semble suggérer une conclusion inverse. Ce cas illustre la problématique de l'absence de prise en compte des corrélations entre les variables par PDP. Pour en tenir compte, il existe une alternative au PDP appelée ICE. Cette méthode est détaillée en annexe \ref{ICE_methodo}.\\

\paragraph{Une variante : l'IPD}
En plus de donner l'effet marginal moyen d'une variable, les graphiques de PDP peuvent fournir une information sur l'importance d'une variable dans la prédiction faite par un modèle. En effet, lorsque le graphe de PDP associé à la variable $X_1$ (par exemple) est relativement plat, il semble naturel de penser que cette variable n'a pas beaucoup d'influence sur la prédiction de $Y$. L'idée introduite par \cayNP{pdp_feature_imp} est ainsi de définir une fonction $Flat$ qui mesure la "platitude" de la courbe de PDP : pour une observation $x, \ i(x)=Flat(\hat{f}_{x_S}(x_S))$.
Une mesure simple et efficace proposée \cay{pdp_feature_imp} est la variance empirique lorsque les variables $x_S$ sont continues et la statistique d'intervalle divisée par 4 pour les variables catégorielles à $K \in \mathds{N}$ niveaux.
Dans le cas où $S=\{1\}$, on obtient alors les formules :
$$i(x_1)=\left\{\begin{matrix}
\frac{1}{n-1}\sum\limits_{i=1}^n[\hat{f}_{x_1}(x_1^{(i)})-\frac{1}{n}\sum\limits_{i=1}^n{\hat{f}_{x_1}(x_1^{(i)})}]^2 \ si \ x_1 \ est \ continue\\ 
[\underset{1 \leq i \leq n}{max} \hat{f}_{x_1}(x_1^{(i)})-\underset{1 \leq i \leq n}{min} \hat{f}_{x_1}(x_1^{(i)})]/4 \ si \ x_1 \ est \ qualitative
\end{matrix}\right.$$
Cette technique est appelée $IPD$ (Importance Based On Partial Dependance). En outre, le graphique PDD permet de fournir une meilleure interprétation des relations entre la variable à expliquer par l'algorithme et les variables endogènes de la base de données.\\

\paragraph{Conclusion : apports et limites du PDP}
Ainsi, le graphique PDP est souvent utilisé pour sa simplicité d'interprétation et sa facilité d'implémentation. De plus, ce graphique peut servir d'outil dans l'estimation de l'importance des variables et leurs interactions au sein d'un modèle. Cependant, ce graphique seul ne suffit pas à expliquer toute la complexité d'un algorithme. La méthode de calcul repose effectivement sur une hypothèse forte et limitante d'indépendance entre les variables. Par ailleurs, le PDP masque les effets hétérogènes comme illustré sur la figure \ref{fig:scatterPDP}. C'est la raison pour laquelle cette méthode est souvent associée à d'autres graphiques comme ICE détaillé en annexe \ref{ICE_methodo}.

\subsubsection{Graphique des effets locaux accumulés (ALE)}

Le graphique des effets locaux accumulés (Accumulated Local Effects Plot) a pour objectif de corriger les limites de ceux de PDP, notamment lorsque les variables explicatives sont corrélées. La méthode a été introduite par \cayNP{ale_pdp}. Tout comme le PDP, l'ALE est une approche globale d'interprétation.\\

\begin{figure}[!h]
    \centering
    \includegraphics[width=90mm]{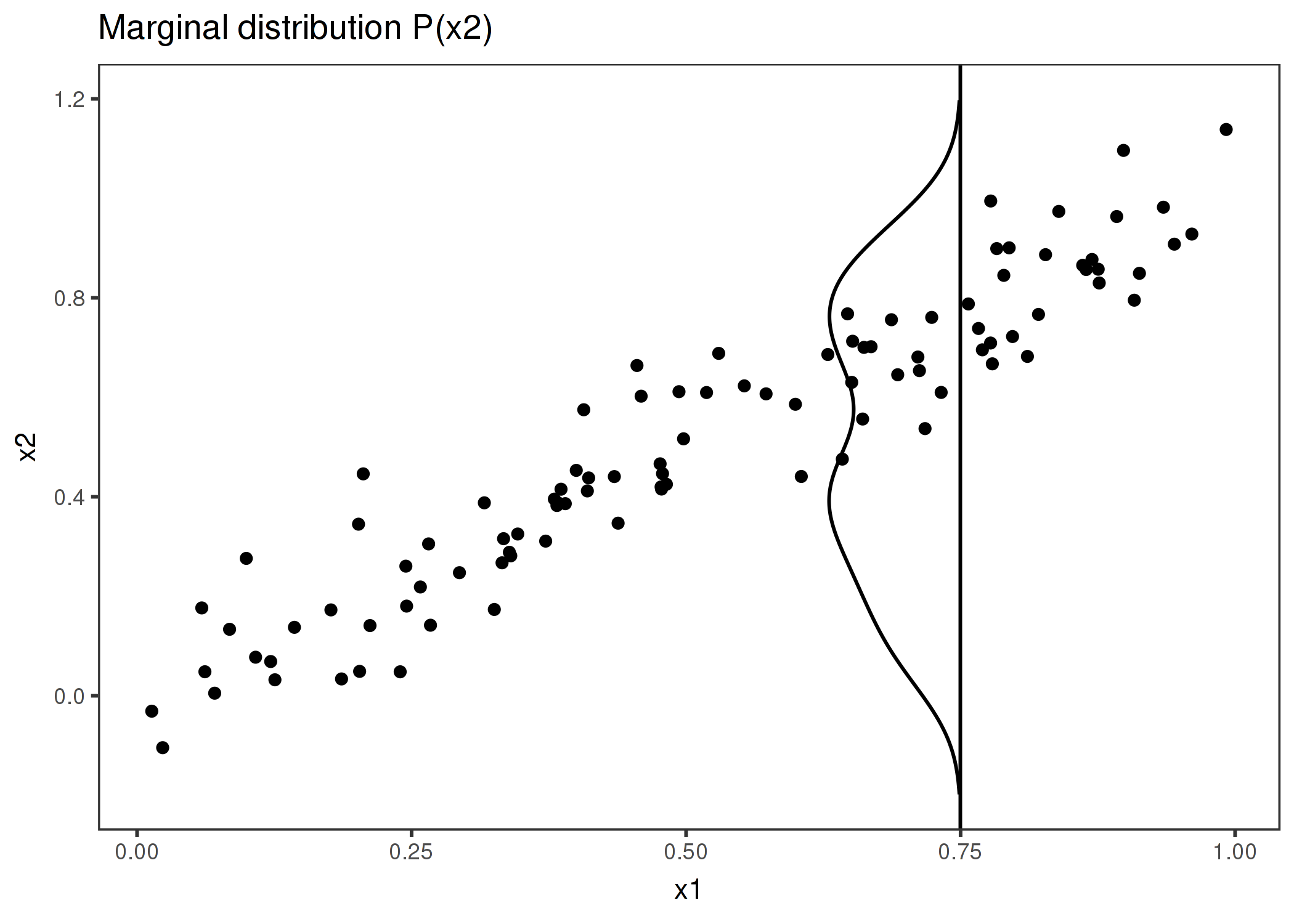}
    \caption{Cas du calcul de la PDP avec des variables très corrélées lorsque l'on fixe $x_1$=0.75 \cay{molnar2019}}
    \label{fig:ale_pdp_pb}
\end{figure}

Partant de l'exemple des combinaisons de poids et tailles introduit par \cayNP{molnar2019}, le PDP illustré en figure \ref{fig:ale_pdp_pb} ne tient effectivement pas compte de la distribution empirique.

\begin{figure}[!h]
    \centering
    \includegraphics[width=90mm]{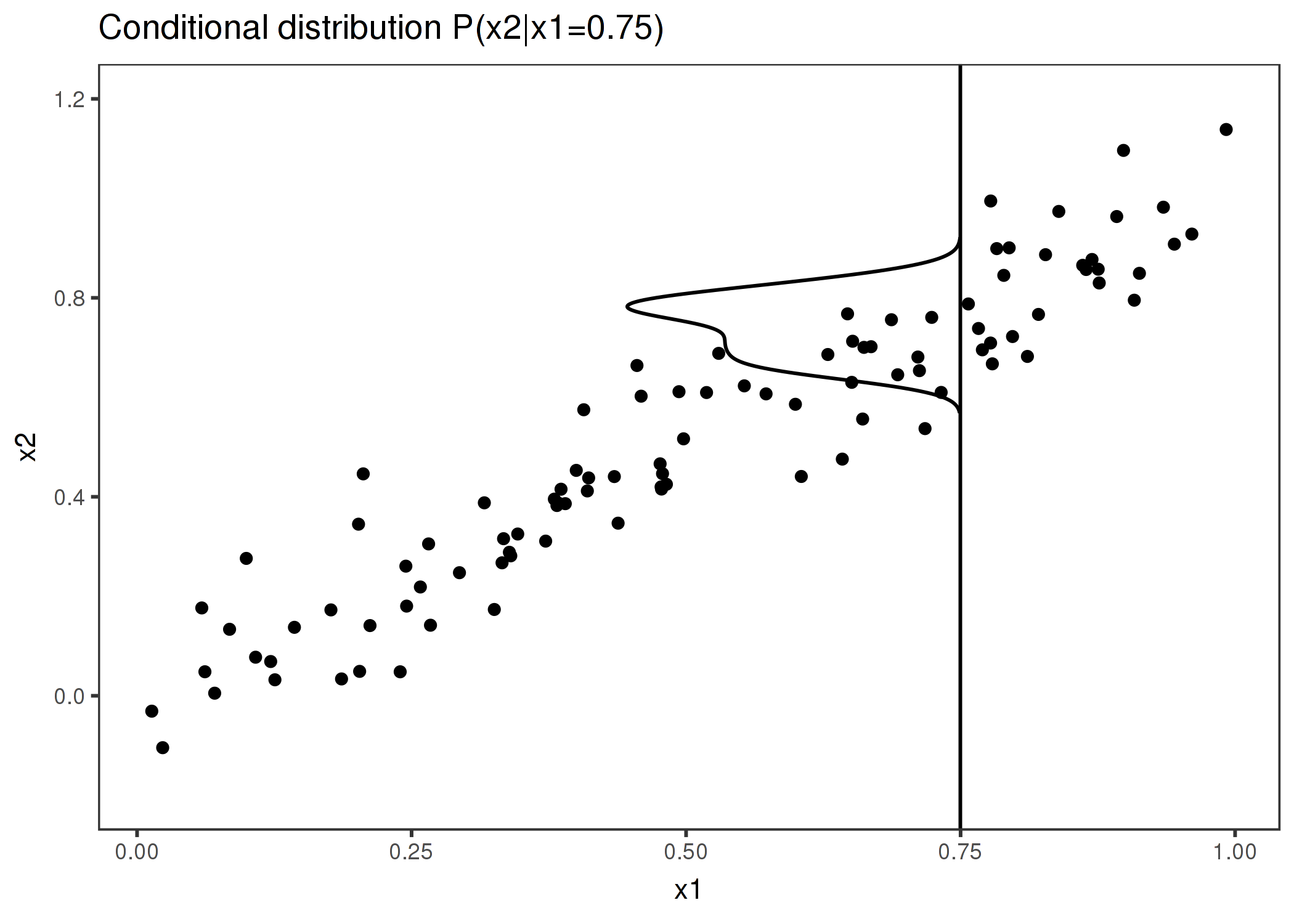}
    \caption{$M$-plot dans le cas de deux variables très corrélées en utilisant la distribution conditionnelle de $x_2$ sachant $x_1=0.75$
    \cay{molnar2019}}
    \label{fig:m_plot}
\end{figure}

Une première idée illustrée en figure \ref{fig:m_plot} afin d'éviter ce problème est, dans le calcul de PDP, de moyenner à partir de la distribution conditionnelle, ce qui signifie que pour une valeur $x_1$ donnée, on réalise la moyenne des instances avec des valeurs similaires localement à $x_1$. 

Cependant cette approche ne permet pas de tenir compte d'effets combinés de variables. Par exemple si on souhaite prédire $Y$ le prix d'une maison, à partir des variables $X_1$ (surface de la maison) et $X_2$ (nombre de chambres), en supposant que la variable de surface de la maison n'a pas d'effet sur la prédiction mais que seul le nombre de chambres en a un, comme le nombre de chambres augmente avec la surface, le $M$-plot précédent montrera alors que la surface de la maison fait augmenter son prix.\\

\begin{figure}[!h]
    \centering
    \includegraphics[width=90mm]{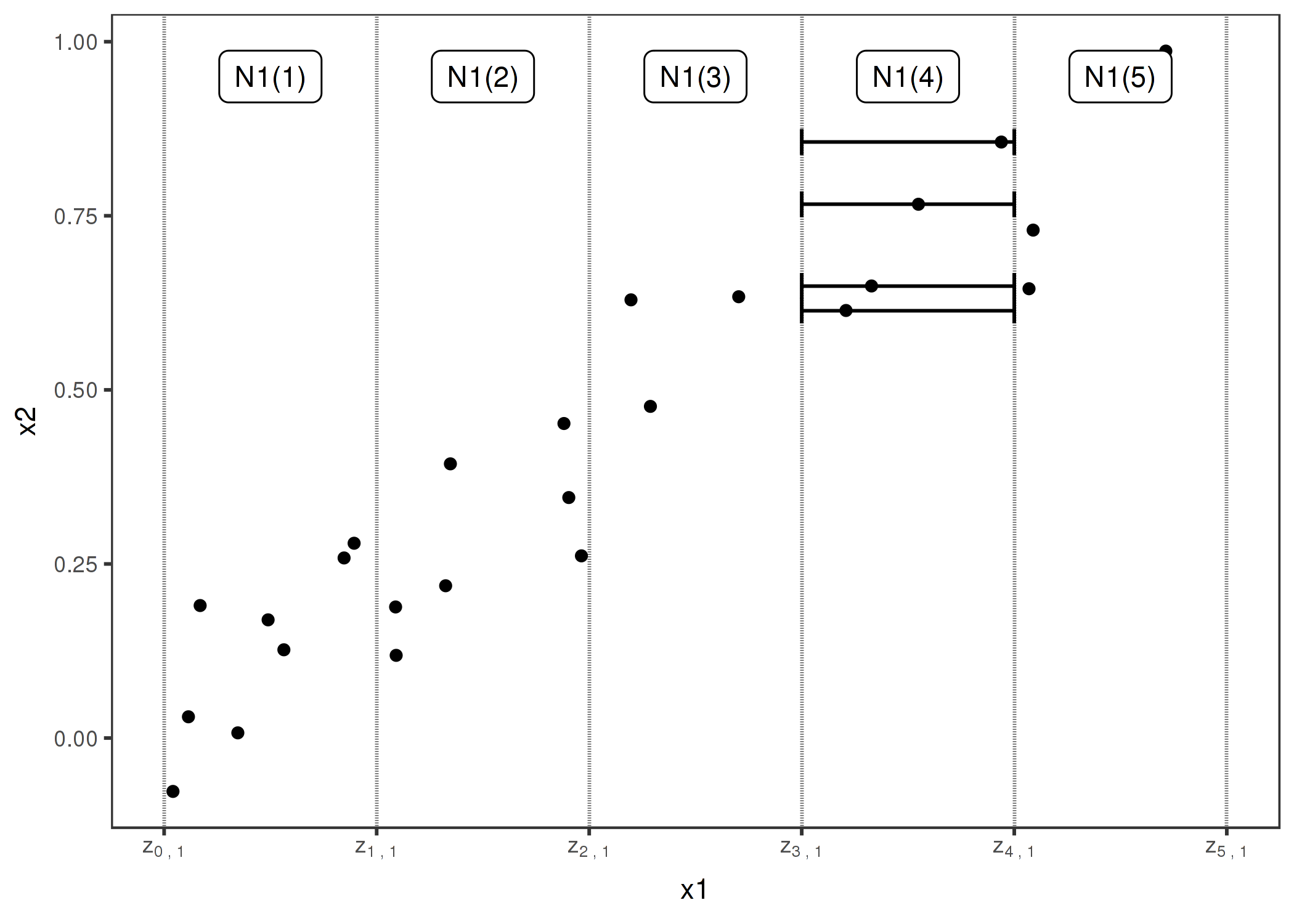}
    \caption{Explication du calcul de l'ALE avec des variables $X_1$ et $X_2$ très corrélées \cay{molnar2019}}
    \label{fig:ale}
\end{figure}

Le graphique des effets locaux accumulés (ALE) dépasse cette limite. Ce dernier repose sur la distribution conditionnelle des variables mais calcule également les différences en prédiction à la place de moyennes. Ainsi si on veut comprendre l'effet associé à une surface de 30 m$^2$, la méthode ALE utilise toutes les instances (ie toutes les maisons) de 30m$^2$ et regarde la différence en prédiction lorsqu'on change leur surface de 29m$^2$ à 31m$^2$. Ceci donne alors l'effet de la variable de surface, et non l'effet combiné avec le nombre de chambres, qui lui est corrélé, comme dans le $M$-plot. Le graphique \ref{fig:ale} résume l'idée de calcul de l'ALE : on divise tout d'abord la variable $X_1$ en intervalles, pour chaque instance dans un intervalle, on calcule la différence en prédiction lorsqu'on remplace la valeur de $X_1$ par la borne supérieure et inférieure de l'intervalle considéré; enfin, toutes ces différences sont accumulées et centrées, ce qui donne la courbe d'ALE. La méthode de construction de la courbe est précisée en annexe \ref{ALE_methodo}. 

Nous venons ainsi d'illustrer deux méthodes qui permettent d'analyser de manière globale tout modèle. Cependant, lorsque les modèles sont complexes ou pas suffisamment parcimonieux, ces outils ne sont parfois pas suffisants pour comprendre les prédictions. Les deux parties suivantes introduisent deux algorithmes récemment publiés proposant des méthodes d'interprétation locales : LIME \cay{lime} et SHAP \cay{shap}.

\subsection{LIME}

\paragraph{Présentation}
LIME \cay{lime} est l'une des premières approches locales apparues dans le domaine du machine learning interprétable. Cette méthode consiste à utiliser un modèle de substitution (noté $M_2$) qui approche localement le modèle que l'on tente d'expliquer  (noté $M_1$). \\

Cette substitution s'effectue en appliquant dans un premier temps une légère perturbation des données initiales $X$. On crée alors un nouvel échantillon, noté $\tilde{X}$. Sur ce dernier, on applique alors le modèle $M_1$ afin de reconstruire la variable à expliquer correspondante. Ainsi, on notera $M_1$: $\hat{y_1}=f_1(\hat{X})$. Chaque observation de l'échantillon $\tilde{X}$ simulé est ensuite pondérée en fonction de sa proximité avec les données initiales : plus celle-ci est proche, plus son poids est important. Sur ces données pondérées, on construit alors un modèle simple d'interprétation $M_2$. Celui-ci est généralement de type Lasso pour la régression et un arbre de décision pour la classification.
Notons que cette fois-ci le modèle $M_2$ fournit une bonne approximation locale mais pas nécessairement une approximation globale.\\

Ainsi la fonction $\hat{g}$ associée au modèle $M_2$ est trouvée en résolvant le problème d'optimisation :  \begin{equation} 
\hat{g}= \underset{g \in G}{argmin} [J(f,g,\pi_x) + \Omega(g)] 
\label{eq:LIME}
\end{equation}

avec $J$ la fonction de coût; $f$ la fonction associée au modèle $M_1$, $g$ la fonction associée au modèle $M_2$ qu'on souhaite optimiser, appartenant à la classe de modèle $G$; $\pi_x$ une mesure de proximité définissant la taille du voisinage autour de $x$ que nous considérons pour l'interprétation du modèle et $\Omega$ une fonction traduisant la complexité d'un modèle.\\

Toutefois en pratique, l'implémentation en Python de LIME n'optimise que le terme associé à la fonction de coût. Il revient à l'utilisateur de choisir un modèle peu complexe, comme par exemple si  $f_1$ est une régression, un modèle avec un nombre limité de variables explicatives (c.f critère de parcimonie).\\

\begin{figure}[!h]
    \centering
    \includegraphics[width=70mm]{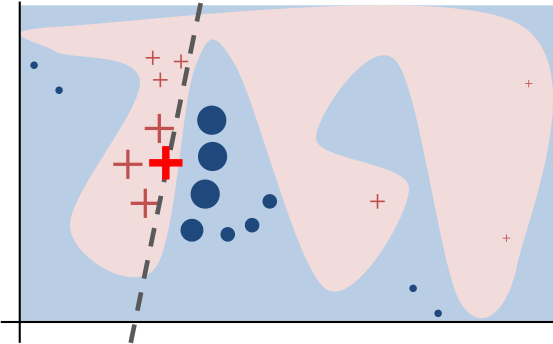}
    \caption{Principe de LIME pour un problème de classification binaire \cay{lime}}
    \label{fig:LIME}
\end{figure}

La figure \ref{fig:LIME} \cay{lime}, résume le fonctionnement de LIME dans le cas d'un modèle de classification binaire (classe 0 ou 1), avec deux variables explicatives. La zone en bleu représente les points associés à la classe 1 selon le modèle étudié et la zone rose clair les points associés à la classe 0. Les croix roses et les points bleus représentent quant à eux les données simulées pour l'apprentissage du modèle de substitution. La taille du motif représente le poids du point considéré, suivant sa distance à l'observation d'intérêt, représentée par la croix rouge. La droite grise en pointillés est la limite de décision obtenue par l'algorithme LIME à l'aide d'un modèle linéaire.

\paragraph{Limites}
LIME a néanmoins fait l'objet de plusieurs critiques. Ces critiques sont de deux ordres principalement.

\begin{figure}[!h]
    \centering
    \includegraphics[width=70mm]{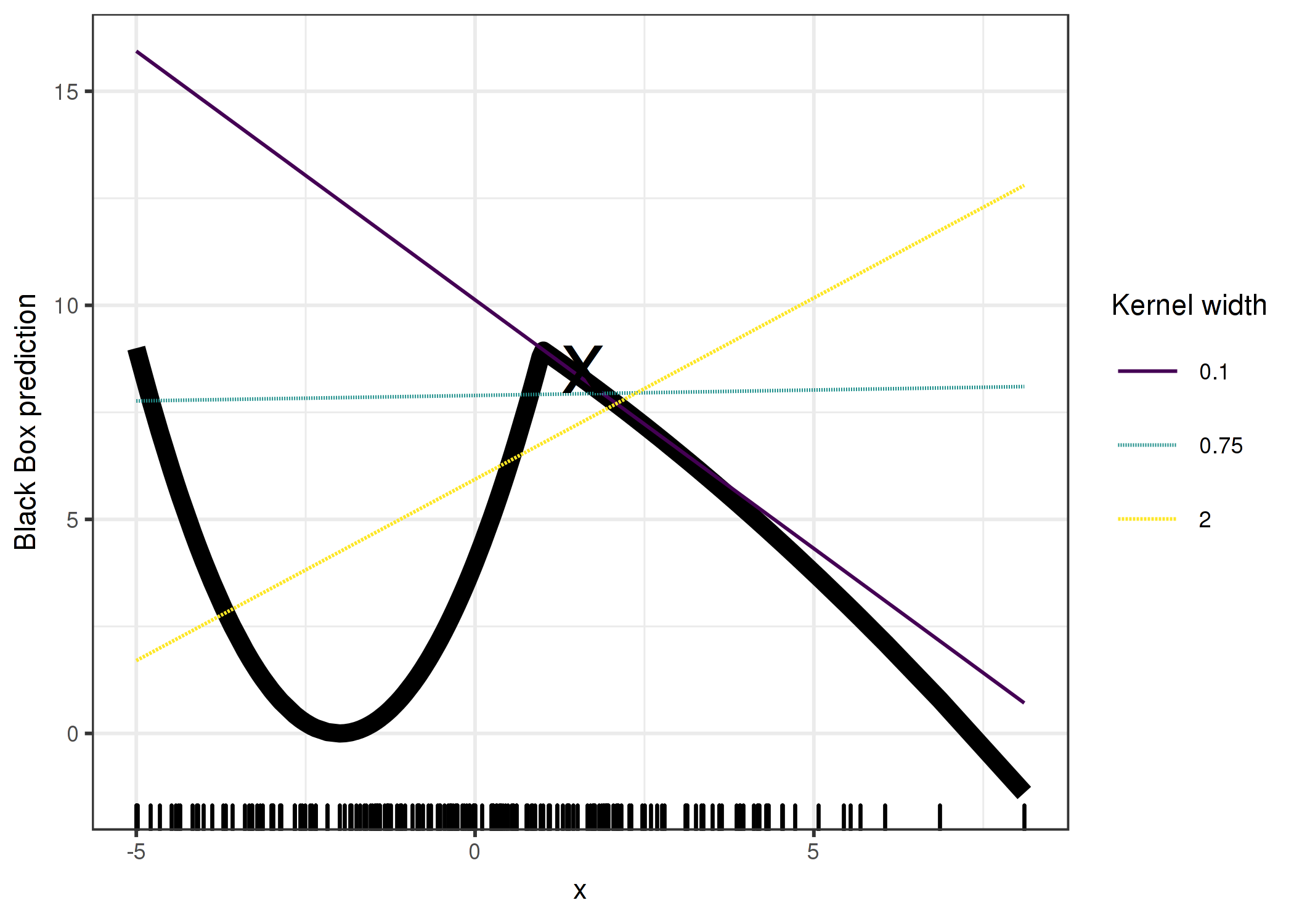}
    \caption{Choix du paramètre du noyau pour la mesure de proximité dans l'algorithme LIME \cay{molnar2019}}
    \label{fig:limite_LIME}
\end{figure}

Tout d'abord, comme le remarque \cayNP{limite_lime}, le choix du noyau utilisé dans l'algorithme LIME pour mesurer la proximité des observations est primordial. Il peut en effet avoir un impact majeur sur la fidélité et la précision de l'explication qui en découle. Par exemple, considérons une variable explicative $X$ et un modèle de décision représenté par le trait noir sur la figure \ref{fig:limite_LIME}. Notre objectif est de comprendre localement la prédiction faite par le modèle au niveau de l'instance $x=1.6$ (représentée par la croix noire). Les lignes tracées de différentes couleurs (jaune, vert et violet) illustrent la sensibilité de l'approximation locale de LIME au paramètre $\sigma$ du noyau. On observe effectivement que les lignes jaunes ($\sigma$=2) et vertes ($\sigma$=0.75) répliquent peu le comportement local du modèle contrairement à la courbe violette ($\sigma$=0.1). L'annexe \ref{alternatives_LIME} détaille une alternative de LIME proposée par \cayNP{BreakDownLive}.

Par ailleurs, \cayNP{why_trust_LIME} soulignent les incertitudes liées aux méthodes d'interprétation locale des modèles de machine learning, dont LIME. Les auteurs mettent en garde les utilisateurs sur la robustesse et la confiance que l'on peut avoir en la méthode. L'utilisation de LIME déplace alors la question de l'interprétabilité des modèles complexes sur les outils eux-mêmes utilisés pour la résoudre.
\cayNP{why_trust_LIME} soulignent notamment via un exemple  : 
\begin{itemize}
    \item \emph{Le hasard dû à l'échantillonnage des données} : comme l'échantillonnage est aléatoire, deux tirages ne donnent pas nécessairement la même explication d'une prédiction
    \item \emph{La sensibilité des explications au choix des paramètres}, comme la taille de l'échantillon et la proximité d'échantillonnage.
    \item \emph{La variation de la crédibilité d'interprétation selon les points étudiés}.
\end{itemize}

\paragraph{Conclusion}
Même si LIME présente certaines limites, cette méthode propose un premier outil d'interprétation locale et permet notamment de compléter les limites des méthodes standards telles que l'importance des variables dans les méthodes par arbres (Random Forest, Gradient Boosting, etc.).

\subsection{SHAP}

SHAP \cay{shap} est également un algorithme d'interprétation locale. Il s'appuie sur la mesure de Shapley introduite en théorie des jeux en 1953.

\subsubsection{La valeur de Shapley en théorie des jeux}

Quand un modèle réalise une prédiction,  nous percevons intuitivement que chaque variable ne joue pas le même rôle : certaines n'ont quasiment aucun impact sur la décision prise par le modèle, alors que d'autres ont beaucoup plus d'influence. L'objectif de l'algorithme SHAP est de quantifier le rôle de chaque variable dans la décision finale du modèle. Pour cela, l'algorithme s'appuie sur la valeur de Shapley \cay{winter2002shapley} - introduisons dans un premier temps cette valeur.\\

Considérons un jeu $J$ caractérisé par un 2-uplet : $J=(P,v)$ où $P=(\{1,...,p\} $ est un ensemble de $p$ joueurs, $p \in \mathds{N}^*$ et $v: S(P) \rightarrow \mathds{R}$ est une fonction caractéristique telle que : $v(\emptyset)=0$  avec $S(P)$ l'ensemble des sous-ensembles de P.\\

Un sous-ensemble de joueurs $S \in S(P)$ est appelé coalition et l'ensemble $ \{1,...,p\}$ de tous les joueurs est appelé la grande coalition. La fonction caractéristique $v$ décrit l'importance de chaque coalition.\\

L'objectif du jeu est alors de répartir l'importance de chaque joueur dans le gain total de la manière la plus "juste" possible. Ainsi, on cherche un opérateur $\phi$, qui assigne au jeu $J=(\{1,...,p\},v)$, un vecteur $\phi=(\phi_1,...,\phi_p)$ de payoffs. Comment définir la notion de répartition juste entre les joueurs ? Lloyd Shapley en propose en 1953 une définition en quatre axiomes :

\begin{itemize}
    \item Efficacité:
    $\sum\limits_{i=1}^p{\phi_i(v)}=v(\{1,...,p \})$
    \item Symétrie:
    Pour tout couple de joueurs $(i,j) \in \{1,...,p \}^2$, si $\forall S \in S(\{1,...,p \}\backslash{\{i,j\}})$, $v(S \cup {i})=v(S \cup {j})$, alors $\phi_i(v)=\phi_j(v)$ 
    \item Facticité:
    Soit $i \in \{1,...,p \} $ un joueur. Si $\forall S \in S(\{1,...,p \}\backslash{\{i\}}), v(S \cup \{i \}) = v(S)$, alors: $\phi_i(v)=0$.
    \item Additivité:
    Pour tous jeux, $v: S(P) \rightarrow \mathds{R}$, $w: S(P) \rightarrow \mathds{R}, \ \phi(v+w)=\phi(v)+\phi(w)$ avec: $\forall S \in S(\{1,...p \}), (v+w)(S)=v(S)+w(S)$
\end{itemize}

La valeur de Shapley $\phi$ est alors l'unique valeur "juste" qui distribue le gain total $v(\{1,...,p\})$, c'est-à-dire celle qui respecte les quatre conditions précédentes. Shapley démontre ce théorème et donne une valeur explicite de cette valeur, à savoir:
\begin{equation} \forall i \in \{1,...,p\}, \ \phi_i(v)=\sum\limits_{S \in S(\{1,...,p\}) \backslash{\{i\}}}{\frac{(p-|S|-1)! |S|!}{p!} (v(S \cup \{i\})-v(S))}
\label{Shapley}
\end{equation}
On peut également définir cette valeur de Shapley d'une autre manière:
\begin{equation} \forall i \in \{1,...,p\}, \ \phi_i(v)=\frac{1}{p!} \sum\limits_{O \in Perm(P)}{v(Pre^i(O) \cup \{i\} ) - v(Pre^i(O))} \end{equation}
avec: $Perm(P)$ l'ensemble des permutations de $P= \{1,...,p \}$ et $Pre^i(O)$ l'ensemble des joueurs qui sont prédécesseurs du joueur $i$ dans la permutation $O \in Perm(P)$ (il s'agit du nombre qui apparaît avant le nombre $i$ dans la permutation $O$).

\subsubsection{La valeur de Shapley appliquée à l'interprétabilité des modèles}

SHAP \cay{shap} reprend la valeur de Shapley pour en faire une mesure du poids de chaque variable dans les prédictions d'un modèle, et ce quelle que soit sa complexité. \\

Considérons une variable numérique à prédire $Y \in \mathds{R}$, à partir d'un vecteur $X \in \mathds{R}^p$ de $p \in \mathds{N}$ variables explicatives.\\

On suppose que l'on dispose d'un échantillon: $y=(y_1,...,y_n) \in \mathds{R}^n$ correspondant aux valeurs cibles et $x=(x_{ij})_{1 \leq i \leq n, 1 \leq j \leq p}$ correspondant aux variables explicatives (avec $n \in \mathds{N}$ le nombre d'individus).\\

Notre algorithme $M$ de machine learning est calibré sur cet échantillon et on note $\hat{g}$ la fonction associée au modèle, c'est-à-dire la fonction qui renvoie la prédiction $\hat{y}$ de y faite par le modèle à partir du vecteur $x$: $\hat{y}=\hat{g}(x)$.\\

Si on fait l'analogie avec la version de la mesure de Shapley en théorie des jeux, nous obtenons :
\begin{itemize}
    \item le jeu: la tâche de prédiction pour une instance $\tilde{x} \in \mathds{R}^p$ du dataset,
    \item le gain: la prédiction actuelle de cette instance moins la prédiction moyenne de toutes les instances du jeu de données,
    \item les joueurs: les valeurs des caractéristiques $x_j, \ j \in \{1,...,p\}$, qui collaborent pour recevoir le gain (ici il s'agit de prédire une certaine valeur).
\end{itemize}

Supposons que notre variable $Y$ à expliquer est le prix d'une voiture en euros et que nos variables explicatives sont $x_1$ et $x_2$, respectivement le nombre de chevaux de la voiture et le nombre de portes. Supposons également que pour $x_1=150$ et $x_2=4$, le prix estimé par le modèle $\hat{g}$ est $\hat{y}=150 000$. Nous savons également qu'à partir des données initiales (constituées de plusieurs prix de voitures et des variables explicatives associées), la prédiction moyenne est de $170 000$ euros.\\

L'objectif du jeu est alors d'expliquer cette différence de $-20 000$ euros, entre la prédiction faite par le modèle et la prédiction moyenne. On pourrait par exemple, obtenir le résultat suivant: $x_1$ a contribué pour $+ 10 000$ euros et $x_2$ pour $- 30 000$ euros (par rapport à la valeur moyenne prédite) et justifierait donc la différence de $- 20 000$ euros observée.\\

Finalement, on peut définir la valeur de Shapley comme la contribution marginale moyenne d'une variable (explicative) sur toutes les coalitions possibles. 

\subsubsection{Un cas particulier : la valeur de Shapley dans le cas de la régression linéaire}

On considère le modèle linéaire: $\hat{g}(x)=\beta_0 + \sum\limits_{i=1}^{p}{\beta_i \ x_i}$, avec $(\beta_i)_{0 \leq i \leq p} \in \mathds{R}^p$.
On définit alors la valeur de Shapley de la variable $j \in {1,...,p}$ associée à la prédiction $\hat{g}(x)$:
$\phi_j(\hat{g})=\beta_j  x_j - \mathds{E}[\beta_j X_j]=\beta_j(x_j - \mathds{E}[X_j])$ (avec $\mathds{E}[\beta_j X_j]$ l'effet moyen de la variable $x_j$).
On parle aussi de contribution de la variable $x_j$ dans la prédiction de $\hat{g}(x)$, car il s'agit de la différence entre l'effet de la variable et l'effet moyen.
On peut remarquer que la somme des contributions de toutes les variables explicatives donnent la différence entre la valeur prédite pour $x$ et la valeur de prédiction moyenne. En effet:
\begin{equation}
    \sum\limits_{j=1}^{p}{\phi_j(\hat{f})}=\sum\limits_{j=1}^p{(\beta_j x_j - \mathds{E}[\beta_j X_j])}= (\beta_0 + \sum\limits_{j=1}^p{\beta_j x_j}) - (\beta_0 +\sum\limits_{j=1}^p{\mathds{E}[\beta_j X_j] })=\hat{g}(x) - \mathds{E}[\hat{g}(X)]
    \label{eq:shap_reg_lin}
\end{equation}
Cette écriture peut alors être généralisée à tout modèle à l'aide de la valeur de Shapley.

\subsubsection{La valeur de Shapley dans le cas général}

Considérons une variable numérique à prédire $Y \in \mathds{R}$, à partir d'un vecteur $X \in \mathds{R}^p$ de $p$ variables explicatives.
On se place dans le cadre d'un modèle quelconque, avec $\hat{g}$ la fonction associée. 
Soit $\tilde{x}=(\tilde{x_1},...,\tilde{x_p})$ l'instance pour laquelle on veut expliquer la prédiction. \\

Définissons la différence en prédiction d'un sous-ensemble des valeurs des caractéristiques dans une instance particulière $\tilde{x}$, introduite par \cayNP{shap_fast}. Il s'agit du changement dans la prédiction causé par l'observation de ces valeurs des variables explicatives.
Formellement, soit $S=\{i_{1},...,i_{s}\} \subset \{1,...,p\}$ un sous-ensemble des variables explicatives (avec $s \in \{1,...,p\}$). Notons $\Delta^{\tilde{x}}$ la différence de prédiction, associée au sous-ensemble $S$:
$$\Delta^{\tilde{x}}(S)=\mathds{E}[\hat{g}(X_1,...,X_p)| X_{i_1}=x_{i_1},...,X_{i_s}=x_{i_s}]-\mathds{E}[\hat{g}(X_1,...,X_p)]$$
Cette différence de prédiction correspond à notre fonction de coût.
Ainsi $(\{1,...,p\}, \Delta^{\tilde{x}})$ forme un jeu de coalition tel qu'il est défini dans la partie précédente.\\

La contribution de la variable explicative $x_j,\ j \in \{1,...,p\}$, est définie comme la valeur de Shapley de ce jeu de coopération $(\{1,...,p\}, \Delta^{\tilde{x}})$: 
\begin{equation}
\phi_j(\Delta^{\tilde{x}})=\sum\limits_{S \in S(\{x_1,...,x_p \}\backslash{ \{x_j\}})}{\frac{|S|! (p-|S-1|)!}{p!}(\Delta^{\tilde{x}}(S \cup \{x_j \}) -\Delta^{\tilde{x}}(S))}
\label{eq:SHAP_2}
\end{equation}
Dans cette formule: $S(\{x_1,...,x_p \}\backslash{ \{x_j\}})$ l'ensemble des permutations de cet ensemble.
En utilisant la formule alternative équivalente, on a également:

\begin{equation}
\forall i \in \{1,...,p\}, \ \phi_i(\Delta^{\tilde{x}})=\frac{1}{p!} \sum\limits_{O \in Perm(P)}{\Delta^{\tilde{x}}(Pre^i(O) \cup \{i\} ) - \Delta^{\tilde{x}}(Pre^i(O))}
\label{eq:shap_3}
\end{equation} 
où  $Perm(P)$ est l'ensemble des permutations de $P=\{1,...,p\}.$

Prenons un exemple simple pour comprendre comment la valeur de Shapley fonctionne. Considérons un jeu avec trois joueurs $\{1,2,3\}$. On compte alors $2^3=8$ sous-ensembles $S$ possibles, à savoir : $\emptyset, \{1\}, \{2\}, \{3\}, \{1,2\}, \{1,3\}, \{2,3\}$ et $\{1,2,3\}$. En utilisant la formule de l'équation \ref{eq:shap_3}, on obtient : 
\begin{align*}
    & \phi_1 = \frac{1}{3}(v(\{1,2,3\}) - \{2,3\}) + \frac{1}{6}(v(\{1,2\}) - \{2\})+ \frac{1}{6}(v(\{1,3\}) - \{3\}) + \frac{1}{3}(v(\{1\})-v(\emptyset) ) \\
   & \phi_2 = \frac{1}{3}(v(\{1,2,3\}) - \{1,3\}) + \frac{1}{6}(v(\{1,2\}) - \{1\})+ \frac{1}{6}(v(\{2,3\}) - \{3\}) + \frac{1}{3}(v(\{2\})-v(\emptyset) ) \\
   & \phi_3 = \frac{1}{3}(v(\{1,2,3\}) - \{1,2\}) + \frac{1}{6}(v(\{1,3\}) - \{1\})+ \frac{1}{6}(v(\{2,3\}) - \{2\}) + \frac{1}{3}(v(\{3\})-v(\emptyset) ) 
\end{align*}
En définissant le gain "non distribué" $\phi_0 = v(\emptyset)$, qui correspond au payoff fixé qui n'est pas associé aux actions des joueurs, la propriété d'additivité est bien respectée, à savoir : $\phi_0 + \phi_1 + \phi_2 + \phi_3 = v(\{1,2,3\})$.

Dans le cas général, on retrouve alors les propriétés vues précédemment à savoir:
\begin{itemize}
    \item Efficacité: $\sum\limits_{i=1}^{p}{\phi_i(\Delta^{\tilde{x}}})=\Delta^{\tilde{x}}(\{1,...,p\})= \hat{g}(\tilde{x})-\mathds{E}[\hat{g}(X)]$.
    On retrouve alors la propriété que l'on a observée pour le modèle linéaire, à savoir que la somme des p contributions pour l'explication d'une observation est égale à la différence entre la prédiction faite par le modèle pour cette observation et la prédiction (moyenne) du modèle si on ne connaissait aucune information sur la valeur des variables explicatives $x_j, \ j \in \{1,...,p\}$.
    \item Symétrie: deux variables explicatives qui ont une influence identique sur la prédiction auront des valeurs de contributions identiques.
    \item Facticité: une variable qui a une contribution de 0 n'aura aucune influence sur la prédiction.
    \item Additivité: Si le modèle qu'on utilise repose sur la moyenne de plusieurs modèles (comme les forêts aléatoires qui utilisent des arbres de décision) alors la contribution de ce modèle sera la moyenne des contributions de chaque modèle pris seul.
    
\end{itemize}
\subsubsection{Algorithme de calcul approché de la valeur de Shapley}

Le problème, en pratique, est le temps de calcul de la valeur de Shapley du fait de sa complexité (croissante avec le nombre de variables et de modalités). En effet, pour ce faire, nous devons calculer toutes les coalitions possibles avec ou sans la variable que l'on souhaite expliquer: la complexité est donc exponentielle. \\

Pour remédier à ce problème, \cayNP{shap_fast} proposent une approximation qui s'appuie sur des méthodes de simulation par Monte Carlo, à savoir:
$$\hat{\phi_j}=\frac{1}{M}\sum\limits_{m=1}^{M}({\hat{g}(x_{+j}^m)- \hat{g}(x_{-j}^m})),$$
où: $j\in \{1,...,p\}$ est l'indice de la variable que nous souhaitons expliquer, $M \in \mathds{N}$ est le nombre d'itérations choisi et $\hat{g}(x_{+j}^m)$ est la prédiction pour le vecteur $x=(x_1,...,x_p)$ de $p$ variables explicatives, mais avec un nombre aléatoire de caractéristiques remplacées par un point $z$ aléatoire, excepté pour la valeur de la caractéristique $j$ choisie. La prédiction
$\hat{g}(x_{-j}^m)$ est quasiment identique à $\hat{g}(x_{+j}^m)$ sauf que la valeur $x_j^m$ est aussi prise à partir de l'échantillon de $x$.\\

On en déduit la procédure proposée par Strumbelj et Kononenko pour approcher la valeur de Shapley $\phi_j(\Delta^{\tilde{x}})$ associée à la variable $x_j$ pour $j \in \{1,...,p\}$ à l'aide de l'algorithme suivant :

\begin{itemize}
    \item Entrée: le modèle $\hat{g}$, l'instance $\tilde{x}$ que nous cherchons à expliquer et $M$ le nombre d'itérations de l'algorithme
    \item $\phi_j=0$
    
\item pour $i$ allant de 1 à $M$, faire:
\begin{itemize}
    \item choisir une permutation aléatoire $O \in Perm(P)$
    \item choisir une instance $z=(z_1,...,z_p)$ du dataset initial
    \item $\ \forall k \in \{1,...,p\}$, $x^+_k=\left\{\begin{matrix}
\tilde{x}_k$  si $k \in Pre^i(O) \cup \{j\}\\ 
z_k$ sinon
$\end{matrix}\right.$

\item $\ \forall k \in \{1,...,p\},x^-_k=\left\{\begin{matrix}
\tilde{x}_k$  si $k \in Pre^i(O)\\ 
z_k$ sinon
$\end{matrix}\right. $
\item $\phi_j=\phi_j+(\hat{g}(x^+)-\hat{g}(x^-))$
\end{itemize}
\item Sortie: $\hat{\phi_j}^{(M)}=\frac{\phi_M}{M}$
\end{itemize}

Notons bien qu'à chaque itération, les calculs des termes $\hat{g}(x^+)$ et $\hat{g}(x^-)$ reposent sur des observations qui sont identiques à l'exception de la variable $\tilde{x}_j$. Ils sont construits en prenant l'instance $z$ et en changeant la valeur de chaque variable apparaissant avant la $j$-ième variable dans l'ordre de la permutation $O$ (pour $x^-$ la valeur de $\tilde{x}_j$ est également changée) par la valeur des caractéristiques de l'instance pour laquelle on désire expliquer $y$.

\subsubsection{Propriétés et limites de SHAP}

Bien que SHAP soit également un modèle d'interprétation local, il diffère de LIME : SHAP explique la différence entre une prédiction et la prédiction moyenne globale, tandis que LIME explique la différence entre une prédiction et une prédiction moyenne locale.\\

SHAP est la seule méthode d'interprétabilité, à ce jour, avec un fondement mathématique. En effet, la différence entre la prédiction et la prédiction moyenne est distribuée de manière "juste" entre les différentes variables utilisées par le modèle, grâce à la propriété d'efficacité de la valeur de Shapley. Ceci n'est pas le cas de LIME, qui repose sur un principe qui semble cohérent mais n'a pas de justification mathématique. SHAP pourrait ainsi être une méthode d'interprétabilité des modèles répondant aux exigences du "droit à l'explication" instauré par le RGPD.\\

La méthode SHAP fournit une explication de la prédiction faite par un modèle quelconque (aussi complexe soit-il) en attribuant une valeur de contribution à chaque variable utilisée, contrairement à LIME qui renvoie une réponse plus concise, en pénalisant les modèles complexes. On peut alors considérer que SHAP réalise moins d'approximations que LIME et de ce fait fournit une explication plus précise. \\

Lorsque le modèle à interpréter est entraîné avec un grand nombre de variables, l'interprétation fournie par SHAP n'est pas parcimonieuse. SHAP renvoie effectivement autant de coefficients que de variables explicatives, ce qui rend parfois la lecture difficile.\\

Pour contourner ce problème, une adaptation de SHAP, appelée \textit{Kernel Shap} (Linear LIME + Shapley Values) est proposée \cay{shap}. L'idée est ainsi de relier les équations \ref{eq:LIME} (de LIME) et \ref{eq:SHAP_2} (de SHAP). En choisissant judicieusement la fonction de coût $J$, la mesure de proximité $\Pi_{x'}$ et le terme de régularisation $\Omega$, il est alors possible d'écrire la valeur de Shapley comme solution du problème d'optimisation posé par LIME dans l'équation \ref{eq:LIME}. Cette combinaison permet alors de fournir des explications plus parcimonieuses.\\

Par ailleurs, SHAP dans sa version initiale suppose que les variables sont indépendantes. Une alternative a néanmoins été récemment proposée par \cayNP{SHAP_dependant}. \\

Enfin, remarquons que SHAP ne fournit qu'une indication sur la contribution de chaque variable pour une prédiction donnée. Il ne permet pas de déduire des effets globaux, contrairement à l'interprétation des \textit{odds-ratios} dans le cadre de régression linéaire. Il n'apporte qu'une compréhension locale, même si cette dernière est parfois plus explicite lors de l'usage de modèles complexes comme les réseaux de neurones ou des méthodes ensemblistes (forêts aléatoires, XGBoost par exemple).

\subsection{Mesure de l'interaction entre les variables à l'aide de la $H$-statistique}
\subsubsection{Principe de l'interaction entre les variables}
L'interaction entre les variables (\textit{feature interaction}) apparaît lorsque les prédictions ne sont pas seulement composées de la somme des effets individuels de chaque variable, mais aussi de termes supplémentaires, correspondant au fait que la valeur d'une variable dépend également de la valeur de l'autre variable.
C'est par exemple le cas lorsque nous mettons en place un modèle de régression linéaire "avec interaction":
\begin{itemize}
    \item $Y=\beta_1 X_1 +\beta_2 X_2 + \varepsilon$ est sans interaction entre $X_1$ et $X_2$
    \item $Y=\beta_1 X_1 +\beta_2 X_2 + \beta_{1,2}X_1 X_2 + \varepsilon$ possède une interaction entre les variables explicatives $X_1$ et $X_2$
\end{itemize}

Considérons un autre exemple, dans lequel nous souhaitons prédire le coût moyen d'un sinistre automobile d'un assuré à partir de son âge (jeune ou vieux) et la puissance de sa voiture (faible ou élevée).
Nous disposons des prédictions suivantes:
\begin{table}[!h]
\begin{center}
\scalebox{0.8}{
\begin{tabular}{|l|c|r|}
  \hline
  Age & Puissance & Prédiction (coût moyen des sinistres)\\
  \hline
  Jeune & Elevée & 300 \\
  Jeune & Faible & 200 \\
  Vieux & Elevée & 250 \\
  Vieux & Faible & 150\\
  \hline
\end{tabular}
}
\caption{Tableau de prédiction du modèle 1, sans interaction}
\label{table:2}
\end{center}
\end{table}

Sur ce modèle très simple, nous pouvons décomposer la prédiction du modèle de la manière suivante: 
\begin{itemize}
    \item un terme constant (\textit{intercept}) de 150
    \item un terme d'effet de l'âge du conducteur de 50 (0 si il est vieux, + 50 si il est jeune)
    \item un terme d'effet de la puissance du véhicule de 100 (0 si le conducteur est âgé, + 100 si il est jeune)
\end{itemize}
Nous n'observons donc pas de terme d'interaction.
Considérons un autre exemple où les prédictions sont les suivantes:
\begin{table}[!h]
\begin{center}
\scalebox{0.8}{
\begin{tabular}{|l|c|r|}
  \hline
  Age & Puissance & Prédiction (coût moyen des sinistres)\\
  \hline
  Jeune & Elevée & 400 \\
  Jeune & Faible & 200 \\
  Vieux & Elevée & 250 \\
  Vieux & Faible & 150\\
  \hline
\end{tabular}
}
\caption{Tableau de prédiction du modèle 2, avec interaction}
\label{table:1}
\end{center}
\end{table}

Sur ce nouveau modèle nous pouvons décomposer la prédiction de cette manière:
\begin{itemize}
    \item un terme constant (\textit{intercept}) de 150
    \item un terme d'effet de l'âge du conducteur de 50 (0 si il est vieux, + 50 si il est jeune)
    \item un terme d'effet de la puissance du véhicule de 100 (0 si le conducteur est âgé, + 100 si il est jeune)
    \item un terme d'interaction entre la variable d'âge et de puissance de 100( +100 si l'assuré est à la fois âgé et possède une voiture puissante, 0 sinon) 
\end{itemize}

\`A l'aide de la $H$-statistique, nous pouvons mesurer l'interaction entre les variables pour n'importe quel modèle \cay{H_stat}.

\subsubsection{$H$-statistique de Friedman}
\label{subsec:Hstat}
Nous utilisons les mêmes notations que pour la partie \ref{subsec:PDP} plus haut sur le PDP, à savoir: $X_S$ représente le sous-ensemble de variables dont nous souhaitons mesurer l'influence, $X_C$ le reste des variables ($C=\{1,...,n\} \backslash{} S$) et $\hat{f}$ le modèle, supposé complexe, que nous étudions. Pour tout $j\in \{1,...,p\}$,  notons $PD_j$ la fonction de dépendance associée à la variable $X_j$ et $PD_{-j}$ la fonction de dépendance associée à toutes les variables sauf $X_j$.  Notons également, pour $j,k\in \{1,...,p\}$, $PD_{j,k}$ la fonction de dépendance associée aux variables $X_j$ et $X_k$.
Rappelons que nous estimons la fonction de dépendance à l'aide de la relation:
\begin{equation}
PD_S(x_S)=\mathds{E}_{X_c}[\hat{f}(x_S,X_C)]\simeq\frac{1}{n}\sum\limits_{i=1}^n{\hat{f}(x_S,x_C^{(i)})}
\end{equation}

Nous supposons dans cette section que le modèle est centré, i.e. 
$\mathds{E}[\hat{f}(X)]=0$.
Dans le cas d'absence d'interaction entre les variables $X_j$ et $X_k$, nous avons alors la relation:
\begin{equation}
    PD_{j,k}(x_j,x_k)=PD_j(x_j)+PD_j(x_k)
\end{equation}
Si $X_j$ n'a d'interaction avec aucune des autres variables, la prédiction du modèle d'une entrée $x$ vérifie donc : 
\begin{equation}
    \hat{f}(x)=PD_j(x_j)+PD_{-j}(x_{-j})
\end{equation}

Les coefficients introduits par Friedman exploitent cette relation pour donner une mesure d'interaction.
Le premier coefficient, noté $H_{j,k}$, mesure la quantité de variance expliquée par l'interaction entre $X_j$ et $X_k$:
\begin{equation}
    H^2_{j,k}=\frac{\sum_{i=1}^n\left[PD_{j,k}(x_{j}^{(i)},x_k^{(i)})-PD_j(x_j^{(i)})-PD_k(x_{k}^{(i)})\right]^2}{\sum_{i=1}^n{PD}^2_{j,k}(x_j^{(i)},x_k^{(i)})}
\end{equation}
S'il n'y a pas d'interaction entre $X_j$ et $X_k$, la $H$-statistique vaut zéro, tandis que si toute la variance de $PD_{j,k}$ est expliquée par la somme des fonctions de dépendance individuelle alors elle vaut 1.

Une deuxième statistique a été introduite par Friedman pour mesurer l'effet d'une variable avec toutes les autres :
\begin{equation}
    H^2_{j}=\frac{\sum_{i=1}^n\left[\hat{f}(x^{(i)})-PD_j(x_j^{(i)})-PD_{-j}(x_{-j}^{(i)})\right]^2}{\sum_{i=1}^n\hat{f}^2(x^{(i)})}
\end{equation}

La $H$-statistique est une mesure relativement intuitive des interactions, elle est toutefois relativement longue à calculer. Lorsque le volume de données est important, elle peut même devenir impossible à calculer. On peut alors sous-échantillonner les données disponibles, mais cela augmente la variance de l'estimation et rend la $H$-statistique instable.

\subsection{Importance des variables}
La notion d'importance des variables dans un modèle a fait l'objet de nombreuses définitions. Certaines d'entres elles sont spécifiques à un modèle ou à une classe de modèles : la $t$-statistique est un exemple dans le cas des modèles linéaires mais il existe également des mesures spécifiques aux modèles à base d'arbres. Ici, nous nous intéressons à une nouvelle définition de l'importance des variables, qui a pour particularité d'être agnostique, indépendante du modèle considéré, notée PFI (\emph{Permutation Feature Importance}) \cay{variable_importance}.

L'idée est de considérer que si une variable est très importante, l'altération de la qualité de ses données perturbera grandement la qualité des prédictions du modèles. Pour cela, on altère artificiellement la qualité des données pour cette variable en permutant toutes ses valeurs. Si la prédiction d'un modèle est grandement modifiée lorsque l'on mélange les valeurs d'une variable, cela signifie que le modèle est sensible aux variations de cette variable et donc qu'elle joue un rôle prépondérant dans le modèle. Inversement, une variable qui pour laquelle une modification de ses valeurs n'impactera que peu la prédiction du modèle ne sera pas considérée comme importante. En résumé, une variable est d'autant plus importante que l'erreur de prédiction du modèle augmente après avoir permuté les valeurs de cette variable considérée.

Décrivons le calcul de cette statistique. Soit $f$ la fonction associée au modèle que l'on souhaite interpréter. On se place toujours dans le cas où nous disposons de $n \in \mathds{N}$ observations: $(x^{(1)},...,x^{(n)}) \in  (\mathds{R}^p)^n$  et $(y^{(1)},...,y^{(n)}) \in \mathds{R}^n$. On note $L$ la fonction d'erreur utilisée, par exemple: $L(y,f(x)) =\frac{1}{2n} \sum\limits_{i=1}^n{(y^{(i)}-f(x^{(i)})^2}$
La procédure pour le calcul de l'importance des $p$ variables du modèle est la suivante :

\begin{itemize}
    \item Calcul de l'erreur d'origine du modèle: $err_1=L(y,f(x))$
    \item pour $j=1$ à $p$ :
    \begin{itemize}
        \item On choisit aléatoirement une permutation $\sigma$ de $\{1,...,n\}$ dans $\{1,...,n\}$.
        On définit une nouvelle matrice $(\tilde{x}_j^{(i)})_{\underset{1 \leq i \leq n}{1 \leq j \leq p}}$ de variables d'entrées, par la formule :
        $$\forall i \in \{1,...,n\}, \forall k \in \{1,...,p\}, \tilde{x}_k^{(i)}= \left\{\begin{matrix}
x_k^{(i)} \ si \ k \neq j \\ 
x_j^{(\sigma(i))} \ si \ k=j
\end{matrix}\right.$$\\
C'est-à-dire que l'on permute (avec $\sigma$) les $n$ observations de la variable $x_j$ et on laisse les autres observations inchangées.
\item On estime l'erreur commise par le modèle sur cette nouvelle matrice d'entrée, à savoir: $err_j=L(y,f(\tilde{x}))$.
\item On calcule l'importance de la variable $x_j$ par la formule: $FI^{(j)}=err_j/err_1$
    \end{itemize}
\item Sortie de l'algorithme : $FI^{(1)},...,FI^{(p)}$, triées par ordre décroissant.
\end{itemize}

Selon ce que le type d'interprétation que l'on souhaite obtenir, il peut être plus pertinent de calculer l'importance des variables sur la base d'apprentissage $B_a$ ou la base de test $B_t$ : le calcul sur $B_a$ permet de savoir à quel point le modèle compte sur chaque variable pour faire une prédiction ; celui sur $B_t$ de savoir à quel point une variable contribue à la performance du modèle sur des données non entraînées.

\subsubsection{Avantages}
Les avantages de cette méthode de mesure de l'importance des variables sont nombreux, on peut citer notamment:
\begin{itemize}
    \item Une mesure intuitive : plus l'erreur est grande quand l'information est détériorée, plus la variable est importante.
    \item Un aperçu global synthétique, comme on pourrait l'avoir avec les coefficients d'un modèle de régression linéaire.
    \item Un critère comparable entre différents modèles.
    \item Une prise en compte à la fois des effets de la variable et de ses interactions avec les autres variables\footnote{Ceci peut également être vu comme un inconvénient.}.
    \item Un calcul efficace qui ne nécessite pas de ré-entraîner le modèle, soit un gain de temps en comparaison avec d'autres méthodes.
\end{itemize}

\subsubsection{Inconvénients}
L'importance par permutation a toutefois quelques inconvénients, à savoir:
\begin{itemize}
    \item Le choix entre les bases d'apprentissage et de test n'est pas très clair.
    \item Le résultat fourni par l'algorithme peut varier grandement du fait du hasard introduit par les permutations.
    \item L'ajout d'une variable corrélée à une autre diminue l'importance de la variable considérée.
    \item Les permutations peuvent fournir des instances irréelles. En effet, lorsque l'on permute une variable au sein d'une instance, on ne fait pas attention au fait que la nouvelle instance soit réellement observable. Ceci est le même problème que celui observé avec le PDP. Considérons par exemple le cas où l'on dispose des variables explicatives de poids et de taille d'un homme. Si on réalise une permutation comme dans l'algorithme ci-dessus, on peut se retrouver avec un individu de taille 2 mètres et de poids 30kg, ce qui n'est pas possible en réalité.
\end{itemize}

Au cours des deux dernières parties, nous avons tout d'abord expliqué l'importance du besoin d'interprétabilité dans l'usage de modèles prédictifs tout en donnant quelques éléments définitionnels à cette notion difficile à cerner, puis dans un second temps nous avons exposé quelques méthodes courantes d'interprétabilité. Appliquons désormais ces méthodes à un cas concret afin d'illustrer leurs apports et leurs limites.

\section{Application des méthodes d'interprétation à la tarification automobile}

Dans cette partie, nous mettons les méthodes d'interprétation en application dans le cadre d'un cas pratique actuariel : la tarification automobile. Nous voulons comprendre si des modèles plus complexes, éventuellement plus performants, peuvent s'interpréter de la même manière que les outils plus classiques utilisés aujourd'hui. Plus précisément, nous souhaitons étudier si ces modèles - souvent dits boîtes-noires- pourraient être réellement déployés tout en respectant les règles imposées par la réglementation (droit à l'explication prévu par le RGPD, contrôles de l'ACPR etc.).

\subsection{Modélisation et comparaison des différentes approches}
\begin{figure}[!h]
    \centering
    \includegraphics[scale = 0.7]{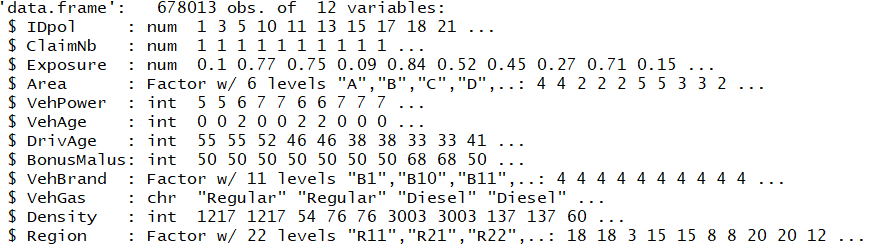}
    \caption{Description des données $freMTPL2freq$ pour modéliser la fréquence des sinistres}
    \label{fig:description_data}
\end{figure}

Nous utilisons la base de données publique \textit{freMTPL2freq}, disponible dans le package R $CASdatasets$.  Il s'agit des données d'un portefeuille français d'assurance de responsabilité civile moteur pour différents assurés observés sur un an. Cette base permet donc de modéliser la fréquence des sinistres. Nous disposons de plus de 600 000 polices d'assurance, avec des variables explicatives comme l'âge de l'assuré, la puissance de véhicule ou encore son ancienneté. La particularité des données de fréquence est qu'elles sont très déséquilibrées, avec de nombreux zéros (absence d'accident) et une exposition variant fortement.
Une description plus détaillée des données est fournie sur la figure \ref{fig:description_data}. 

Dans un contexte de tarification automobile, l'approche actuarielle classique est une modélisation indépendante de la fréquence (nombre de sinistres annuels) et de la sévérité (coût moyen d'un sinistre), pour former la prime pure en combinant les prédictions de ces deux modèles. Nous nous intéressons ici uniquement à la partie fréquentielle. 
Notons que la modélisation de la sévérités s'appuie sur un jeu de données de taille réduite, étant donné qu'uniquement les assurés sinistrés sont utilisés. De plus, la corrélation entre les variables explicatives et la variable cible (montant du sinistre) est généralement assez faible ce qui rend difficile d'obtenir un modèle prédictif performant. Nous avons observé au cours de notre étude qu'il était difficilement possible, y compris avec des modèles complexes tels que les méthodes ensemblistes, d'améliorer sensiblement les performances du GLM Gamma classiquement utilisé.
Concentrons-nous à présent sur le modèle de fréquence.

Avant de mettre en place les algorithmes répondant à ce problème, des traitements préliminaires ont été effectués en tentant de répliquer les pratiques opérationnelles : analyse des valeurs aberrantes et des sinistres extrêmes (qui ont été écrêtés), retraitements des variables catégorielles etc. (voir \cay{Delcaillau} pour plus de détails).
Cette dernière étape est essentielle pour la mise en place d'un modèle linéaire généralisé (GLM) car, sans retraiter les données, une monotonie est imposée pour les variables numériques de par la nature linéaire du modèle. Nous avons repris le retraitement proposé dans l'article \cay{case_study}. 

Notre objectif dans cette partie sera de comparer l'interprétabilité de deux modèles : un modèle GLM classique, très souvent utilisé en actuariat et un autre modèle, plus complexe, donnant éventuellement de meilleures performances. Une fois ces modèles mis en place, nous voulons montrer à l'aide des outils d'interprétation détaillés ci-avant qu'il est possible de comprendre le modèle de boîte-noire implémenté et qu'il n'est pas nécessairement moins interprétable que le modèle GLM.

Dans la gamme des modèles complexes tels que les \textit{Random Forest} ou les réseaux de neurones, nous avons finalement opté pour le modèle eXtrem Gradient Boosting (XGBoost) \cay{xgboost}. Le XGBoost est devenu très populaire dans de nombreuses compétitions de machine learning, comme Kaggle, grâce à la flexibilité permise par ses nombreux hyperparamètres. Nous les avons optimisés à partir de validations croisées. Tout au long de cette partie, nous noterons $A$ le modèle trivial, renvoyant la moyenne de la fréquence des sinistres, $B$ le meilleur modèle GLM (Poisson) trouvé et $C$ le modèle qualifié de boîte-noire, qui est un XGBoost.

\begin{figure}[!h]
    \centering
    \includegraphics[scale = 0.95]{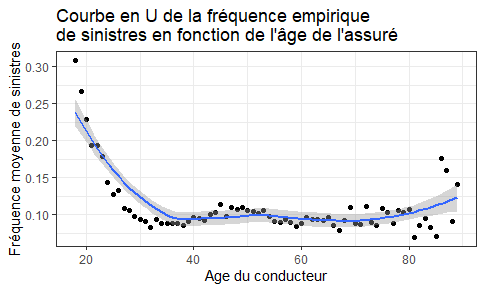}
    \caption{Relation empirique observée entre la fréquence moyenne de sinistres et l'âge du conducteur}
    \label{fig:courbe_en_u}
\end{figure}

Il est important de noter que dans le cas du GLM, il est impératif d'avoir recours à un retraitement  préalable des variables numériques et de les rendre catégorielles. En effet, sans celui-ci, les relations entre les variables explicatives numériques et la sortie seraient toutes monotones. En particulier, la courbe "en U" classiquement observée représentant la relation entre l'âge du conducteur et la fréquence moyenne de sinistres ne pourrait être obtenue sans ce retraitement (cf. figure \ref{fig:courbe_en_u} obtenue avec nos données).
Dans le cadre du XGBoost, et plus généralement des modèles non linéaires, ce retraitement n'est pas nécessaire, car ceux-ci sont capables de capter ces relations complexes.
Afin de pouvoir comparer le GLM et le XGBoost, nous avons completé l'analyse en ajustant un nouveau modèle XGBoost, noté \textit{C-cat}, qui utilise la même transformation des variables que le GLM. Les paramètres de ce modèle ont également été optimisés à l'aide de validation croisée.

Les résultats obtenus pour ces différents modèles sont donnés dans le tableau \ref{tab:res_glm_xgb}. La métrique d'évaluation retenue est ici la déviance de Poisson, très couramment utilisée pour des données fréquentielles. Nous avons également indiqué les valeurs du MSE (\textit{Mean Squared Error}) et du MAE ({\textit Mean Absolute Error}) à titre indicatif. Entre parenthèses sont indiqués les gains relatifs par rapport au modèle trivial. 
Ces critères ont été à la fois calculés sur la base d'apprentissage (\textit{In-Sample}), pour tester la capacité des modèles à s'ajuster aux données d'entraînement, ainsi que sur la base de test (\textit{Out-of-Sample}) pour tester la capacité du modèle à s'adaper à de nouvelles données. On note un gain de l'ordre de 3\% sur la déviance de Poisson du XGBoost sur le meilleur modèle GLM. Dans un contexte de forte concurrence comme l'assurance automobile, ce gain de précision pourrait s'avérer essentiel, notamment pour ne pas récupérer les mauvais risques et ainsi éviter l'anti-sélection. Notons toutefois qu'il convient de nuancer ces propos puisqu'il n'est pas prouvé qu'une forte segmentation soit synonyme de profit dans un milieu concurrentiel. En effet, segmenter, en plus d'être en opposition avec le principe de base de l'assurance, à savoir la mutualisation, conduit à une augmentation de la volatilité des résultats  \cay{segmentationPlanchet}.

\begin{table}[!h]
\begin{tabular}{c|c|c|c|c|c|c|}
\cline{2-7}
\multirow{2}{*}{}                        & \multicolumn{2}{c|}{Déviance de Poisson}                                                                                & \multicolumn{2}{c|}{MSE}                                                                                                & \multicolumn{2}{c|}{MAE}                                                                                                \\ \cline{2-7} 
                                         & App.                                                       & Test                                                       & App.                                                       & Test                                                       & App.                                                       & Test                                                       \\ \hline
\multicolumn{1}{|c|}{Modèle trivial ($A$)} & 32.94                                                      & 33.86                                                      & 0.0564                                                     & 0.0596                                                     & 0.0995                                                     & 0.1015                                                     \\ \hline
\multicolumn{1}{|c|}{Meilleur GLM ($B$)}   & \begin{tabular}[c]{@{}c@{}}31.27\\ \textit{(+5.06\%)}\end{tabular}  & \begin{tabular}[c]{@{}c@{}}32.17\\ \textit{(+4.99\%)}\end{tabular}  & \begin{tabular}[c]{@{}c@{}}0.0557\\ \textit{(+1.28\%)}\end{tabular} & \begin{tabular}[c]{@{}c@{}}0.0589\\ \textit{(+1.66\%)}\end{tabular} & \begin{tabular}[c]{@{}c@{}}0.0979\\ \textit{(+1.62\%)}\end{tabular} & \begin{tabular}[c]{@{}c@{}}0.0999\\ \textit{(+1.60\%)}\end{tabular} \\ \hline
\multicolumn{1}{|c|}{XGBoost ($C$)}        & \begin{tabular}[c]{@{}c@{}}30.22\\ \textit{(+8.24\%)}\end{tabular} & \begin{tabular}[c]{@{}c@{}}31.29\\ \textit{(+7.59\%)}\end{tabular} & \begin{tabular}[c]{@{}c@{}}0.0548\\ \textit{(+2.95\%)}\end{tabular} & \begin{tabular}[c]{@{}c@{}}0.0582\\ \textit{(+2.35\%)}\end{tabular} & \begin{tabular}[c]{@{}c@{}}0.0965\\ \textit{(+3.02\%)}\end{tabular} & \begin{tabular}[c]{@{}c@{}}0.0988\\ \textit{(+2.74\%)}\end{tabular} \\ \hline
\multicolumn{1}{|c|}{XGBoost cat (\textit{C-Cat})}        & \begin{tabular}[c]{@{}c@{}}30.34\\ \textit{(+7.89\%)}\end{tabular} & \begin{tabular}[c]{@{}c@{}}31.37\\ \textit{(+7.36\%)}\end{tabular} & \begin{tabular}[c]{@{}c@{}}0.0549\\ \textit{(+2.71\%)}\end{tabular} & \begin{tabular}[c]{@{}c@{}}0.0582\\ \textit{(+2.23\%)}\end{tabular} & \begin{tabular}[c]{@{}c@{}}0.0966\\ \textit{(+2.87\%)}\end{tabular} & \begin{tabular}[c]{@{}c@{}}0.0988\\ \textit{(+2.68\%)}\end{tabular} \\ \hline
\end{tabular}
\caption{Résultats des différents modèles de fréquence -modèle trivial, GLM, XGBoost, XGBoost cat - basés sur les critères de déviance de Poisson, MSE et MAE}
\label{tab:res_glm_xgb}
\end{table}

Toutefois, au-delà des performances et de l'impact de ce gain de précision sur le résultat de l'assureur, le propos de l'article réside essentiellement dans la capacité d'interpréter les prédictions faites par ce modèle complexe et la possibilité de le comprendre au même titre que le GLM.

\subsection{Interprétation des modèles GLM et XGBoost}
\subsubsection{Le modèle GLM}

\begin{figure}[!h]
    \centering
    \includegraphics[scale = 0.76]{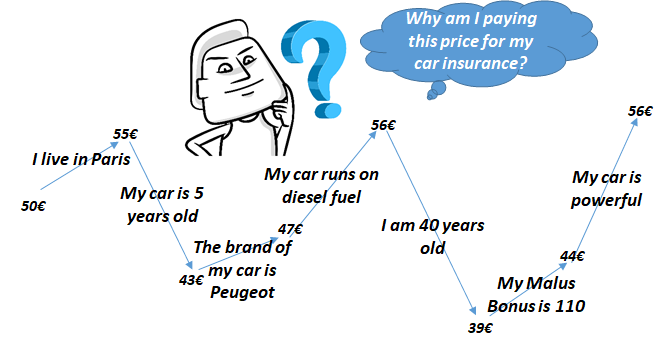}
    \caption{Exemple de justification du prix de l'assurance automobile d'un assuré en fonction de ses caractéristiques}
    \label{fig:prix_glm}
\end{figure}

\paragraph{Un modèle à interprétation intrinsèque}
Une fois les performances des deux modèles étudiées, notamment avec une analyse de la stabilité vis-à-vis de l'échantillonnage de la base de test et d'apprentissage, nous avons mis en pratique les méthodes d'interprétation développées dans la partie \ref{sec:1_methodesInterpretation}, pour mieux comprendre les prédictions réalisées. Nous avons analysé dans un premier temps les prédictions du modèle GLM. Les propriétés de parcimonie et de simulabilité (vues dans la partie \ref{sec:0_interpretabilite}) nous ont permis de comprendre directement le modèle à partir des coefficients de chacune des variables. En particulier, nous avons décomposé le cheminement qui mène à une prédiction pour un assuré donné (cf. figure \ref{fig:prix_glm}).
Il est également aisé d'étudier le changement de prédiction fournie par le GLM lorsque les caractéristiques d'un assuré sont modifiées : il suffit de regarder le changement du coefficient induit par cette modification de caractéristique.

Ainsi, la structure du GLM permet de comprendre les décisions prises par l'algorithme et justifie qu'il soit classé dans la catégorie des modèles intrinsèquement interprétables.

A ce stade, nous disposons d'une compréhension quasi-totale du modèle GLM. C'est pourquoi nous voulons vérifier la cohérence avec les différents outils de la partie \ref{sec:1_methodesInterpretation} pour interpréter tout type de modèles, y compris donc le GLM. 

\begin{figure}[!h]
    \centering
    \includegraphics[scale = 0.8]{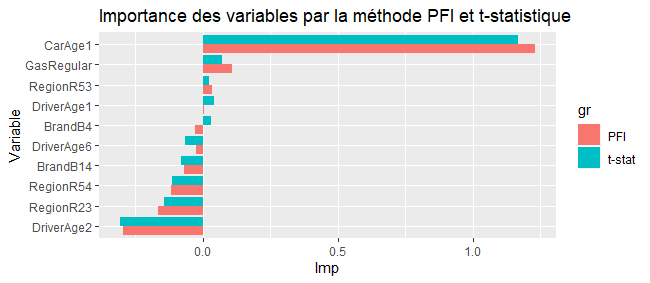}
    \caption{Importance des variables dans le modèle GLM par les approches PFI et $t$-statistique.}
    \label{fig:PFI_vs_TSTAT}
\end{figure}

\paragraph{Interprétation globale. }
Tout d'abord une première question que l'on peut se poser concerne l'importance des variables.
Comme nous l'avons vu dans la partie \ref{sec:1_methodesInterpretation}, la méthode PFI est une approche possible pour y répondre. 
Nous voulons vérifier que les résultats obtenus sont similaires à ceux de la $t$-statistique, propre au GLM, qui repose sur un principe totalement différent.
La figure \ref{fig:PFI_vs_TSTAT} montre les résultats obtenus par ces deux approches, qui sont sensiblement proches. Notons que seules quelques variables sont représentées dans un souci de lisibilité.

\begin{figure}[!h]
    \centering
    \includegraphics[scale = 0.9]{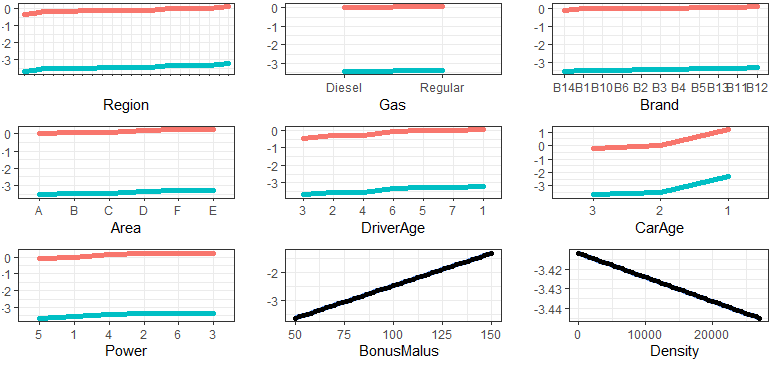}
    \caption{Graphique de dépendance partielle (en rouge) des variables au sein du modèle $B$ (GLM fréquence) et les coefficients du GLM associés (en bleu)}
    \label{fig:pdp_glm}
\end{figure}

L'importance des variables permet de comprendre le rôle global de chaque variable dans la prédiction du modèle GLM mais ne nous indique pas l'impact moyen de chaque modalité sur la prédiction. Pour ce faire, nous pouvons utiliser l'outil de PDP (cf. partie \ref{subsec:PDP}), qui mesure l'effet marginal moyen d'une variable sur la prédiction.
Comme nous pouvons le voir sur la figure \ref{fig:pdp_glm}, les courbes de PDP sont simplement les translations des coefficients du GLM (à la fonction de lien inverse, ici exponentielle, près) pour les variables catégorielles, et les PDP associées aux variables numériques sont des droites. Cela est cohérent avec la théorie du GLM et du PDP et nous conforte dans l'idée que l'information seule des coefficients du modèle suffit à son interprétation. Notons que l'ALE peut également être utilisé et conduit aux mêmes interprétations que le PDP dans le cas du GLM.

\paragraph{Interactions. }
Comme le modèle considéré est un GLM, aucune interaction entre les variables n'est présente. Notons que nous aurions pu utiliser la famille des GAM (\textit{Generalized Additive Model}) pour y inclure des interactions entre certaines variables.

Vérifions que les outils de la section \ref{sec:1_methodesInterpretation} pour identifier les interactions sont en adéquation avec cette analyse. 
Tout d'abord, la $H$-statistique (cf. partie \ref{subsec:Hstat}), fournit des coefficients proche de 0 pour chaque variable, ce qui signifie l'absence d'interaction au sein du modèle.

\begin{figure}[!h]
    \centering
    \includegraphics[scale = 0.8]{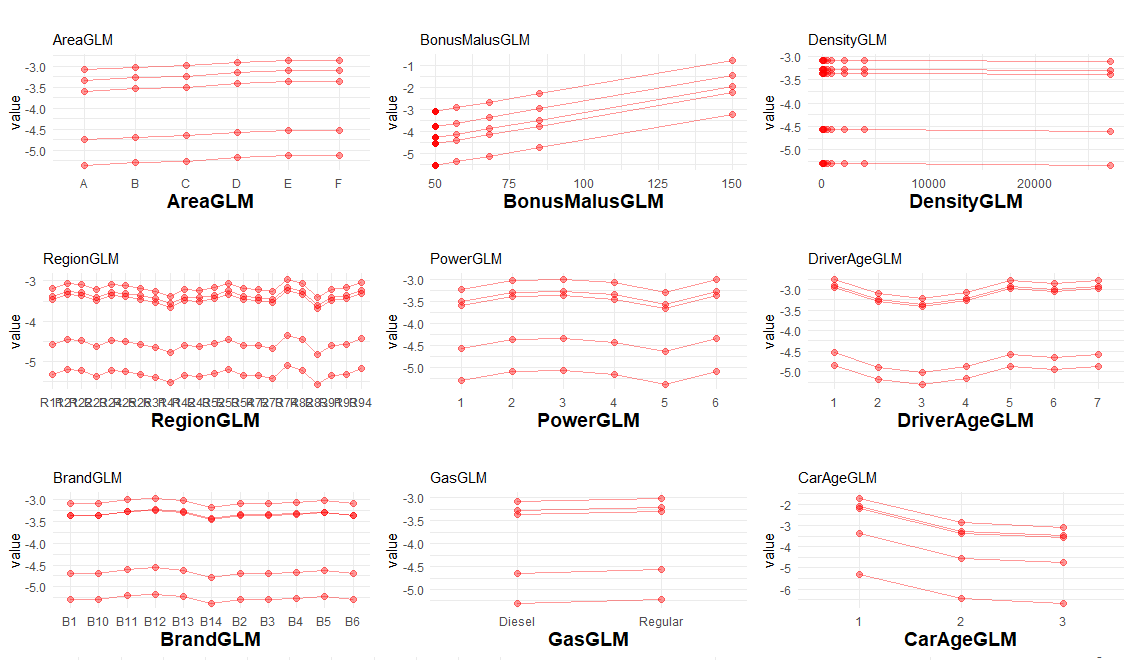}
    \caption{Quelques courbes ICE (à la fonction lien inverse près) associées à chaque variable utilisée dans le modèle GLM}
    \label{fig:ice_glm}
\end{figure}

Nous pouvons également représenter les courbes \textit{ICE} qui permettent d'identifier de possibles interactions entre les variables. Nous observons, sur la figure \ref{fig:ice_glm}, que ces différentes courbes sont translatées entre elles (et avec la courbe PDP qui n'en est que la moyenne), signe d'absence d'interaction.

\paragraph{Interprétation locale. }
Nous pouvons également utiliser les outils d'analyse locale, tels que \textit{LIME} et \textit{SHAP}. Nous observons que les différentes valeurs obtenues pour interpréter localement une prédiction peuvent être directement déduites des coefficients du modèle GLM~: en raison de la linéarité du GLM, l'analyse globale se suffit à elle-même.

\begin{figure}[!h]
    \centering
    \includegraphics[scale = 0.9]{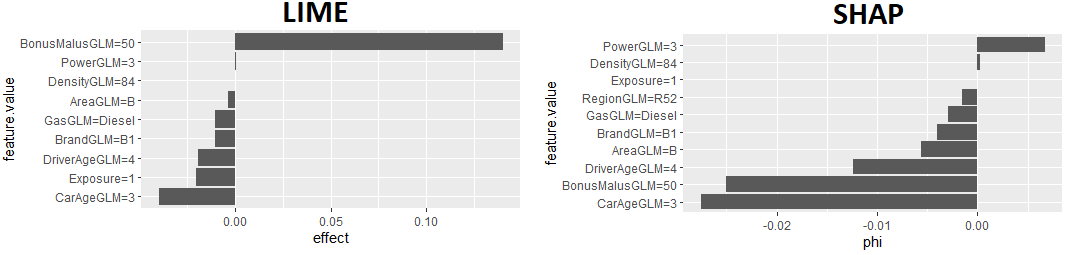}
    \caption{Résultats obtenus via les méthodes d'interprétation locales LIME et SHAP sur un assuré pour le modèle GLM $A$}
    \label{fig:lime_shap_glm_diff}
\end{figure}

Afin d'illustrer concrètement la différence entre LIME et SHAP, il est intéressant d'étudier les résultats de ces deux méthodes dans le cadre du GLM, modèle que l'on maîtrise parfaitement (cf. figure \ref{fig:lime_shap_glm_diff}).

Ainsi, que ce soit pour analyser globalement le comportement du modèle GLM, pour identifier les possibles interactions ou pour comprendre une prédiction au niveau local, les outils du chapitre \ref{sec:1_methodesInterpretation} ne sont pas nécessaires : les informations en sortie du modèle et ses paramètres sont suffisants.

\subsubsection{Le modèle XGBoost}
Nous avons donc montré l'adéquation des résultats théoriques des différentes méthodes d'interprétation agnostiques sur un algorithme parfaitement maîtrisé, à savoir le GLM. Mettons à présent en application ces méthodes sur un modèle considéré comme boîte-noire : le XGBoost. Reprenons pour cela les différentes étapes permettant l'interprétation de notre modèle XGBoost, à savoir l'analyse globale, l'analyse des interactions et enfin l'anayse locale.

\paragraph{Interprétation globale. }

Tout d'abord, concernant l'importance des variables, nous avons recours à la méthode PFI vue précédemment. Il existe également des méthodes intrinsèques au modèle XGBoost, mais nous ne les aborderons pas ici. 

\begin{figure}[!h]
    \centering
    \includegraphics[scale = 0.95]{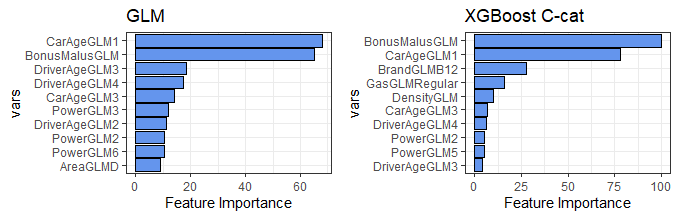}
    \caption{Importance des 10 variables "dummy" les plus importantes dans les modèles GLM (\textit{A}) et XGBoost (\textit{C-cat})}
    \label{fig:imp_var_xgb_glm}
\end{figure}

La figure \ref{fig:imp_var_xgb_glm} donne les scores d'importance des différentes variables utilisées par le XGBoost (et par le GLM). Afin de pouvoir réaliser une comparaison, seul le modèle \textit{C-cat} (XGBoost avec les variables retraitées) est analysé conjointement au GLM.
En effet, la méthode PFI donne un score à chaque variable dite \textit{Dummy} et n'est donc pas directement comparable au modèle XGBoost \textit{C}.

Nous  observons que pour le GLM et le XGBoost, la modalité $1$ de la variable $CarAge$ (i.e voiture âgée de moins d'un an) semble avoir un rôle prépondérant sur les prédictions faites par ces deux modèles. Le même constat est fait pour la variable numérique de bonus malus.

Le fait que la méthode PFI repose sur les modalités de chaque variable, ce qui donne 52 coefficients dans le cas du retraitement des variables réalisé, rend l'analyse difficile et non parcimonieuse. 

\begin{figure}[!h]
    \centering
    \includegraphics[scale = 0.95]{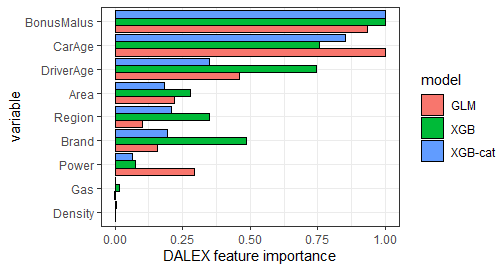}
    \caption{Importance des variables des différents modèles via la méthode \textit{DALEX}}
    \label{fig:imp_var_dalex}
\end{figure}

Le package \textit{DALEX} disponible sour $R$ propose une adaptation de cette méthode d'importance des variables en donnant un score par variable et non plus par modalité.
A l'aide de cette méthode, nous disposons d'une analyse simplifiée de nos différentes variables et de leurs rôles au sein du modèle. Cela nous permettra en outre de comparer nos trois modèles $B$, $C$ et \textit{C-cat}. Le graphique \ref{fig:imp_var_dalex} résume l'importance des variables des trois modèles cités précédemment : globalement le rôle des variables est similaire au sein de chacun des modèles. En particulier, les variables de bonus-malus et d'âge du conducteur sont prépondérantes pour les trois modèles. Nous remarquons également que l'âge du conducteur a un rôle plus important dans le XGBoost ($C$) que dans les autres modèles. Enfin, les variables $Gas$ et $Density$ sont très peu influentes en moyenne sur la prédicition réalisée par les différents modèles. Pour le GLM cela signifie que pour toutes les prédictions faites, quel que soit l'individu concerné, ces variables n'ont pas (ou très peu) d'impact.
 
Dans le cadre des GLM, l’utilisation de méthode d’interprétation locale semble peu pertinente. La seule connaissance des coefficients (pouvant être associés à l’interprétation globale) du modèle suffit à comprendre les comportement locaux. Localement, les coefficients des GLM s’appliquent de manière uniforme à tous les individus afin de produire les prédictions.
Pour le XGBoost, cela signifie que pour la majorité des prédictions faites, ces variables n'auront qu'un impact très faible, mais il se peut, qu'au niveau local (pour quelques individus), ces variables soient prépondérantes. Ce phénomène peut être mis en exergue par les méthodes Lime et Shap notamment.

Nous avons à présent connaissance du rôle global de chaque variable au sein de chaque modèle. Néanmoins, l'importance des variables ne nous fournit pas l'effet moyen de la valeur prise par une variable sur la prédiction du modèle. 
Les méthodes de PDP et ALE permettent de répondre à cette question.

\begin{figure}[!h]
    \centering
    \includegraphics[scale = 0.8]{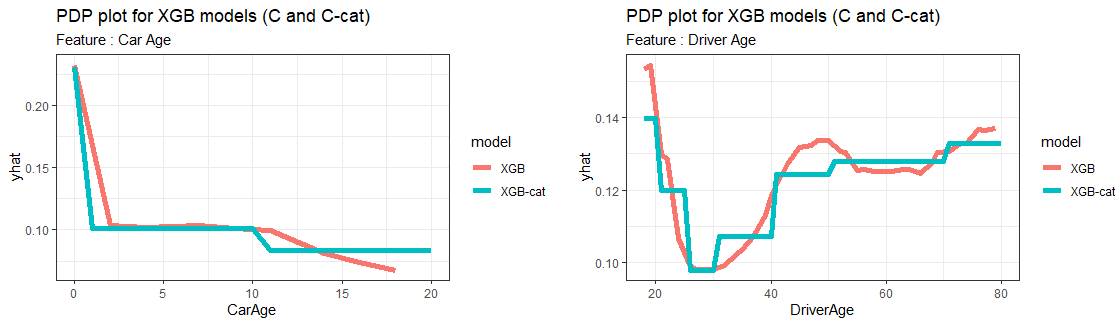}
    \caption{Graphiques de dépendance partielle ($PDP$) associés à la variable d'âge du conducteur et à la variable d'âge du véhicule pour les modèles XGBoost $C$ et \textit{C-cat}}
    \label{fig:pdp_driver_car_age_xgb_xgb_cat}
\end{figure}

Ceci permet, entre autre, de vérifier la cohérence de l'impact de chaque variable sur la sinistralité prédite par rapport à l'analyse empirique réalisée avant modélisation.
En particulier, pour la variable d'âge du conducteur on espère retrouver une courbe en forme de "U", c'est à dire une sinistralité élevée pour les conducteurs jeunes (moins de 25 ans) et âgés (plus de 70 ans), et une sinistralité relativement faible ou modérée pour les âges intermédiaires.
Ceci est bien observé sur le graphique droit de la figure \ref{fig:pdp_driver_car_age_xgb_xgb_cat}. Notons que la courbe bleue est en escalier car les variables ont été discrétisées au préalable. Nous retrouvons d'ailleurs les différentes classes d'âges créées : $[18,21[, [21, 26[, [26,31[, [31,41[, [41, 51[, [51, 71[, [71, +\infty[$.

Nous pouvons réaliser la même étude sur la variable d'âge du véhicule. Sur le graphique gauche de la figure \ref{fig:pdp_driver_car_age_xgb_xgb_cat}, nous observons que plus l'âge de véhicule est faible, plus la fréquence de sinistre est élevé. Autrement dit, la relation antre l'âge de véhicle et la sinistralité est décroissante. Notons que les courbes ont été bornées à un âge de véhicule inférieur à 20 ans. Cette interprétation est en ligne avec celle faite à l'aide du modèle GLM. Afin de vérifier la pertinence des deux modèles, on peut  confirmer ces résultats par une étude empirique sur l'âge du véhicule et la fréquence des sinistres. Une étude descriptive de la base de données permet de vérifier que 30\% des sinistres surviennent pour des véhicules âgés de moins de deux ans. Ce constat peut s'expliquer par la correlation forte entre les grands conducteurs, plus exposés aux sinistres et le fait qu'ils changent plus régulièrement de véhicule. Cependant cet a priori ne peut avec les données actuelles être confirmé. 

En annexe \ref{annexe:graphes} (figures \ref{fig:pdp_all_xgb_freq_num} et \ref{fig:pdp_all_xgb_freq_cat}) sont représentés les graphiques de dépendance partielle pour les autres variables. 
Nous retrouvons notamment, la relation de croissance entre la variable cible et le bonus malus. Les analyses de dépendance partielle permettent donc de retrouver des interprétations similaires à celle des GLM. Combiner par exemple ces outils au Xgboost permettrait par exemple une meilleure classification du risque tout en concervant l'explication des résultats globaux. 

\paragraph{Interactions. }
Cette analyse univariée, à l'aide des graphiques de PDP ou d'ALE, ne montre pas les effets multivariés présents au sein de la base. 
En effet, comme nous étudions un modèle XGBoost, les prédictions ne sont pas aussi simples à expliquer que pour le GLM. Par l'étude du modèle Xgboost seul, la prédiction ne peut pas être exprimée comme la somme des effets individuels de chaque variable car l'effet d'une variable dépend de la valeur des autres variables. L'article \cay{trees_interac} montre que les méthodes construites à partir d'arbres - comme le XGBoost - sont vantées pour leur capacité à modéliser l'interaction entre différentes variables. 
Pour mettre en évidence ces possibles interactions, le calcul de la H-statistique est une solution. Celle-ci, étudiée dans la partie \ref{subsec:Hstat}, estime la force d'interaction en mesurant la part de la variance due à l'effet d'interaction entre plusieurs variables. Son calcul repose en grande partie sur la dépendance partielle, avec un ratio de la variance due à l'interaction et de la variance totale. La valeur de la H-statistique est comprise entre 0 et 1 avec 0 référant à l'absence d'interaction et 1 indiquant que la prédiction est purement guidée par l'interaction étudiée.
Dans notre cas, nous nous restreignons à l'étude conjointe de deux variables. 
Comme la formule de la H-statistique repose sur des dépendances partielles, le temps de calcul est très élevé et des approximations doivent être réalisées.
De plus, dans notre étude seules des variables catégorielles sont utilisées ce qui rend l'utilisation de la H-statistique peu pertinente.

\begin{table}[ht]
\centering
\begin{tabular}{rrrrrrrrrr}
  \hline
 & Area & BonusMalus & Density & Region & Power & DriverAge & Brand & Gas & CarAge \\ 
  \hline
Area &    & 0.002 & 0.002 & 0.001 & 0.002 & 0.001 & 0.001 & 0.003 & 0.006 \\ 
  BonusMalus &    &    & 0.014 & 0.015 & 0.011 & 0.014 & 0.016 & 0.029 & 0.045 \\ 
  Density &    &    &    & 0.013 & 0.011 & 0.006 & 0.007 & 0.009 & 0.063 \\ 
  Region &    &    &    &    & 0.006 & 0.005 & 0.004 & 0.011 & 0.043 \\ 
  Power &    &    &    &    &    & 0.005 & 0.005 & 0.015 & 0.110 \\ 
  DriverAge &    &    &    &    &    &    & 0.005 & 0.011 & 0.021 \\ 
  Brand &    &    &    &    &    &    &    & 0.013 & 0.027 \\ 
  Gas &    &    &    &    &    &    &    &    & 0.067 \\ 
  CarAge &    &    &    &    &    &    &    &    &    \\ 
   \hline
\end{tabular}
\caption{$H$-statistique pour les différents couples de variables utilisés par le XGBoost (obtenu sur 10 000 points, d'où un temps d'exécution important)} 
\end{table}

Ceci vient du fait que la $H$-statistique surestime l'effet des interactions à cause des variables catégorielles, comme c'est le cas avec l'exemple de la combinaison $(Power,Gas)$, ayant respectivement 6 et 2 modalités.
De plus, même si les valeurs de la H-statistique étaient cohérentes, cela ne nous aurait pas donné d'information concernant la nature de l'interaction entre les deux variables choisies.
Pour mieux comprendre comment l'interaction joue un rôle dans les prédictions, nous pouvons par exemple étudier les courbes ICE (\emph{Individual Conditional Expectation} : espérance conditionnelle individuelle), qui généralisent le graphique de dépendance partielle pour chaque observation (c.f partie \ref{subsec:PDP}). A ce titre, la méthode ICE est une méthode d'interprétation locale.

\begin{figure}[!h]
    \centering
    \includegraphics[scale = 0.5]{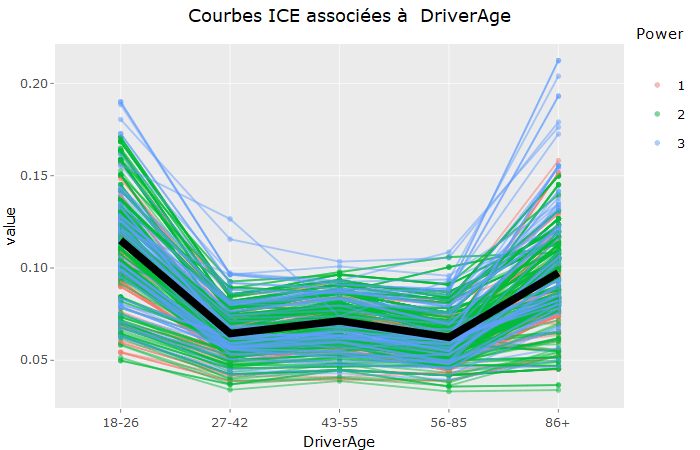}
    \caption{Courbes ICE associées au modèle XGBoost $C$ avec les variables d'âge du conducteur et de puissance du véhicule}
    \label{fig:ice_driver_age_power}
\end{figure}

Les courbes ICE reposent sur un principe simple : si les courbes ne sont pas translatées entre elles (ni avec le graphique de dépendance partielle qui est la moyenne de chaque courbe ICE), cela signifie qu'une interaction est présente entre les variables. 
De par la nature non-linéaire du XGBoost, nous nous attendons évidemment à des courbes non translatées. Afin de mettre en exergue de manière plus précise les possibles interactions, il peut être intéressant de mettre une couleur différente pour chaque courbe suivant la modalité prise par une variable choisie. Nous avons choisi d'étudier l'effet conjoint de l'âge du conducteur et de la puissance du véhicule, répartie ici en 3 classes. Plusieurs études ont montré que l'effet couplé d'un conducteur jeune et d'un véhicule de puissance élevée augmentait drastiquement le risque de sinistres (en terme de fréquence).
Ce phénomène peut-être observé sur la figure \ref{fig:ice_driver_age_power}.
Pour mettre en évidence la remarque précédente sur l'effet combiné de ($DriverAge$, $Power$), il faut comparer les courbes bleues, pour lesquelles la puissance du véhicule est la plus élevée, à la courbe noire de dépendance partielle représentant l'effet moyen de l'âge du conducteur sur la prédiction. On remarque de nombreuses courbes bleues ayant une pentification plus importante entre les modalités $18-26$ et $27-42$, que le PDP, signe d'un risque supplémentaire de sinistres lorsque l'assuré est jeune et qu'il possède une voiture puissante.
La même remarque peut être faite lorsque l'assuré est à la fois vieux (plus de 86 ans) et possède une voiture puissante (de catégorie 3).

\begin{figure}[!h]
    \centering
    \includegraphics[scale = 0.85]{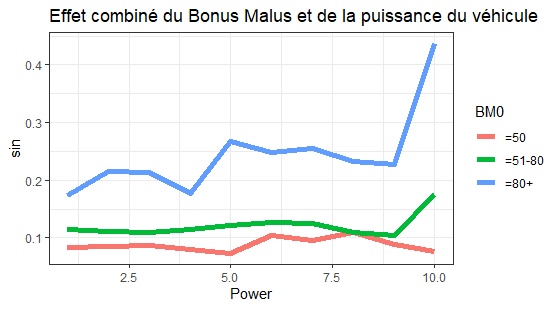}
    \caption{Courbes PDP associées au modèle XGBoost $C$ avec les variables de bonus malus et de puissance du véhicule}
    \label{fig:ice_BM_power}
\end{figure}

Par le biais de ces courbes ICE, d'autres phénomènes peuvent être observés comme l'effet combiné du bonus-malus et de la puissance du véhicule (cf. Figure \ref{fig:ice_BM_power}).

\begin{figure}[!h]
    \centering
    \includegraphics[scale = 0.85]{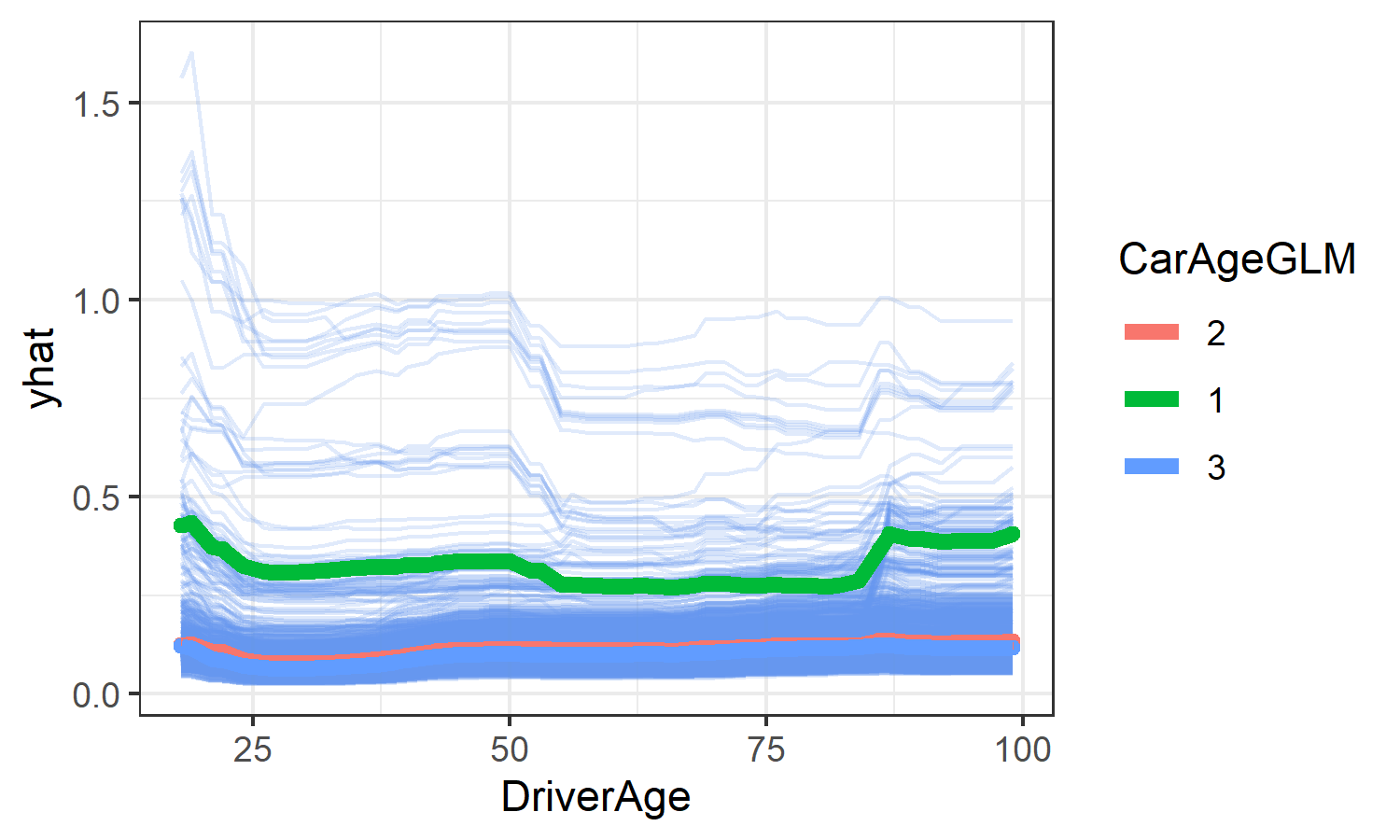}
    \caption{Courbes ICE associées au modèle XGBoost $C$ avec les variables d'âge du conducteur et d'âge du véhicule}
    \label{fig:ice_driver_age_carAge}
\end{figure}

Nous  pouvons également mettre en évidence le risque accru de sinistres lorsque l'assuré est à la fois jeune et possède une voiture récente (de moins d'un an). Le graphique \ref{fig:ice_driver_age_carAge} montre cet effet, sur lequel sont représentés les PDP associés à la variable $DriverAge$ par classe d'âge du véhicule. En vert, il s'agit des véhicules âgés de moins d'un an, en rouge des véhicules âgés entre un et dix ans et enfin en bleu des véhicules de plus de 10 ans. Enfin, les autres courbes bleues (plus fines) correspondent à des courbes ICE issues de $1000$ assurés pris au hasard dans notre base de données. On retrouve une approximation du graphique de dépendance globale associée à la variable $DriverAge$ en calculant la moyenne empirique de ces différentes courbes ICE.

Nous notons sur la courbe verte une décroissance bien plus marquée, pour les jeunes conducteurs, que pour les deux autres courbes. De même, une forte hausse de risque de sinistralité prédite par le modèle $C$ est observée aux alentours des 80 ans pour les véhicules récents.
Notons également que les différentes courbes ICE mettent en exergue des individus atypiques, pour lesquels le nombre de sinistres prédit (sur un an) est très élevé, dépassant même les 1.5 sinistres prédits.
Il peut être intéressant d'étudier en détail ces individus pour lesquels une étude plus approfondie (via les méthodes locales comme LIME et SHAP par exemple) semble indispensable pour comprendre les prédictions du modèle.

\section*{Conclusion}
Les algorithmes prédictifs sont de plus en plus prisés par les industries afin d'améliorer les différents services qu'elles proposent. Parfois moins contraignants en termes d'hypothèses à vérifier que des algorithmes statistiques plus traditionnels, les modèles de machine learning (au sens large: \textit{deep learning}, méthodes ensemblistes, etc.) permettent d'améliorer la compréhension de phénomènes et d'en anticiper la survenance avec une meilleure précision. Ce gain se fait souvent au détriment de la complexité des algorithmes mais également de la donnée sur lesquels ils s'appuient. Les méthodes d'interprétation des algorithmes permettent d'étendre les outils d'analyse de ces modèles afin d'en assurer le contrôle et surtout, la pertinence métier. Pouvoir comprendre et expliquer sont des élements essentiels pour les disciplines fondées sur la manipulation des données. Ainsi, cet article met en avant certaines méthodes d'aide à l'interprétation des algorithmes. Il est important de souligner que la recherche ne cesse d'avancer sur ce sujet tant d'un point de vue technique qu'éthique.

\nocite{*}
%\bibliographystyle{plainnat}   
%\bibliography{bibliographie.bib}

\appendix
\section{Annexe : quelques librairies open source utiles}
\label{libsOpen}

Nous listons ici quelques librairies utiles auxquelles le lecteur peut accéder afin de manipuler les méthodes présentées.

\paragraph{Package R :}
\begin{itemize}
    \item LIME : https://cran.r-project.org/web/packages/lime/index.html;\\ https://cran.r-project.org/web/packages/iml/index.html
    \item SHAP : https://github.com/slundberg/shap 
    \item H-statistique : https://cran.r-project.org/web/packages/iml/index.html
    \item ALE, PDP, ICE : https://cran.r-project.org/web/packages/iml/index.html;\\ https://cran.r-project.org/web/packages/pdp/index.html;\\ https://www.rdocumentation.org/packages/ICEbox/versions/1.1.2
    \item Importance des variables :
    https://cran.r-project.org/web/packages/iml/index.html;\\
    https://cran.r-project.org/web/packages/DALEX/index.html
\end{itemize}

\paragraph{Package Python :}
\begin{itemize}
    \item LIME : https://github.com/marcotcr/lime
    \item SHAP : https://github.com/slundberg/shap 
    \item H-statistique : https://pypi.org/project/sklearn-gbmi/
    \item ALE, PDP, ICE : https://github.com/SauceCat/PDPbox; \\ 
    https://scikit-learn.org/ 
\end{itemize}
\section{Annexe : Algorithme de construction de la courbe PDP}

\label{algo_pdp}
L'algorithme proposé dans \cay{pdp} afin d'estimer les valeurs prises par la fonction de dépendance partielle est le suivant :

\begin{itemize}
    \item Entrée : la base d'apprentissage $(x_j^{(i)})_{1 \leq i \leq n, 1 \leq j \leq p}$, le modèle $\hat{f}$, une variable à expliquer supposée ici être $x_1$  pour simplifier, i.e. $S=\{1\}$ et $C=\{2,...,p\}$. \\
    c'est-à-dire que : $(x_j^{(i)})_{1 \leq i \leq n, 1 \leq j \leq p} = (x_1^{(i)},x_C^{(i)})_{1 \leq i \leq n}$
    \item Pour $i$ allant de $1$ à $n$:
    \begin{itemize}
        \item Copie de la base d'apprentissage, en remplaçant la valeur de la variable $x_1$ par la valeur constante $x_1^{(i)}$: $(x_1^{(i)},x_C^{(k)})_{1 \leq k \leq n}$
        \item Calcul du vecteur de prédiction par $\hat{f}$ des données précédemment définies : $\hat{f}(x_1^{(i)},x_C^{(k)})$ pour $k=1,...,n$
        \item Calcul de $\hat{f}_{x_1}(x_1 ^{(i)})$ par la formule: $\hat{f}_{x_1}(x_1^{(i)}) \simeq \frac{1}{n} \sum\limits_{k=1}^{n}{\hat{f}(x_1^{(i)},x_C^{(k)})} $
    \end{itemize}
    \item Sortie : le graphique des points $(x_1^{(i)},\hat{f}_{x_1}(x_1^{(i)}))$ pour $i=1,...,n$, appelé graphique de dépendance partielle (PDP).
\end{itemize}
\section{Annexe : Précision sur la méthode ICE}
\label{ICE_methodo}
L'approche par les courbes ICE fournit un graphique avec une ligne pour chaque instance, qui montre comment la prédiction change quand une caractéristique change. 
A la place de la moyenne réalisée lors du calcul de la PDP, le calcul de l'ICE est réalisé pour chaque instance. Nous obtenons ainsi un graphique ICE, contenant autant de courbes que d'observations. Cette méthode a été introduite par \cayNP{ice}. Contrairement au graphique de dépendance partielle qui est une approche globale, les courbes ICE sont locales (c.f figure \ref{scope_interpretability}). 
L'algorithme utilisé pour estimer l'ICE est le suivant:

\begin{itemize}
    \item Entrée: les données d'entraînement: $(x^{(i)}_j)_{1 \leq i \leq n,1 \leq j \leq p}$, le modèle ajusté $\hat{f}$, $S$ un sous-ensemble de $\{1,...p\}$ et $C$ le complémentaire de $S$ dans $\{1,...,p\}$. 
    \item Pour i=1,...,n:
    \begin{itemize}
        \item $\hat{f}^{(i)}_S=0_{n\times1}$
        \item $x_C=x_C^{(i)}$: on fixe les colonnes $C$ à la $i$-ième observation.
        \item Pour l=1,...n:
        \begin{itemize}
            \item $x_S=x_S^{(l)}$
            \item $\hat{f}^{(i)}_{S,l}=\hat{f}(x_S,x_C)$
        \end{itemize}
    \end{itemize}
    \item Output: $\hat{f}^{(1)}_{S}=(\hat{f}^{(1)}_{S,1},...,\hat{f}^{(1)}_{S,n}),...,\hat{f}^{(n)}_{S}=(\hat{f}^{(n)}_{S,1},...,\hat{f}^{(n)}_{S,n})$
\end{itemize}

Reprenons l'exemple illustratif de la méthode PDP  page \pageref{fig:scatterPDP}. Dans celui-ci, nous avons observé que la PDP, en réalisant une moyenne, ne capte pas toute la dépendance d'une variable sur la prédiction. 
Affichons à présent le graphique d'ICE associé à la PDP.
\begin{figure}[!h]
    \centering
    \includegraphics[width=80mm]{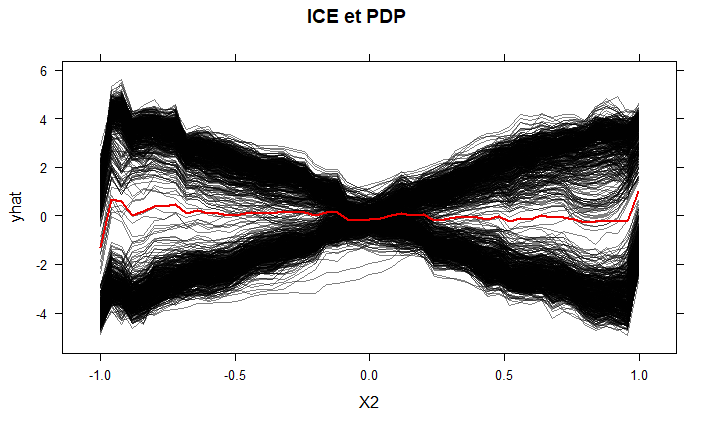}
    \caption{Graphique de PDP (en rouge) et courbes ICE (en noir)}
    \label{fig:ice_pdp}
\end{figure}
Cette fois, l'ICE capte l'effet de $X_2$ sur la prédiction pour chaque instance. En moyenne le graphique d'ICE est proche de 0, ce qui correspond à la PDP, mais le graphique d'ICE complet est proche du nuage de points représenté sur la figure \ref{fig:scatterPDP}.
Il existe une adaptation de la méthode ICE, appelée c-ICE, permettant de centrer les courbes et ainsi de mieux voir les effets de chaque variable sur la prédicition. D'autres variantes  s'appuyant sur les dérivées partielles telles que le d-ICE existent également.\\

Les courbes ICE présentent globalement les mêmes intérêts que le graphique PDP, notamment la simplicité d'implémentation et d'interprétation. L'autre avantage important est qu'elles ne masquent pas les effets hétérogènes du modèle considéré. Ainsi, en utilisant le graphique PDP, qui fournit un résumé de l'impact d'une variable sur la prédiction du modèle, et les courbes ICE, qui le précisent, nous obtenons une bonne explication globale des prédictions.\\

Cependant, tout comme le PDP, les courbes ICE reposent sur une hypothèse d'indépendance entre les différentes variables et ne tiennent pas compte de leur distribution réelle.
Un autre inconvénient est le fait de ne pouvoir représenter ses courbes qu'en 2 ou 3 dimensions, du fait que l'humain ne sait pas se représenter des dimensions supérieures. 
De plus, le graphique contenant toutes les courbes ICE est vite surchargé lorsque le nombre d'individus étudié est grand. 
\section{Annexe : la méthode ALE}
\label{ALE_methodo}
Pour bien comprendre la méthode ALE, détaillée dans \cayNP{ale_pdp}, reprenons la formule de la PDP.
La PDP associée aux variables $X_S$ repose sur le calcul de \begin{equation}\hat{f}_{X_S,PDP}(x_S)=\mathds{E}_{X_C}[\hat{f}(x_S,X_C)]=\int\limits_{x_C}\hat{f}(x_S,x_C)\mathds{P}_{X_c}(x_C)dx_C\end{equation} pour chaque point $x_S$ de la distribution marginale de la variable $X_S$.
Le $M$-plot, lui repose sur le calcul de la moyenne des prédictions sur la distribution conditionnelle, à savoir:
\begin{equation}\hat{f}_{X_S,M-plot}(x_S)=\mathds{E}_{X_C|X_S}[\hat{f}(X_S,X_C)|X_S=x_S]=\int\limits_{x_C}\hat{f}(x_S,x_C)\mathds{P}(x_C|x_S)dx_C\end{equation}
Finalement, pour le graphique ALE, il nous faut définir une borne $z_{0,1}<x_S$ pour délimiter l'intervalle sur lequel on va faire la moyenne des différences de prédiction. Le calcul est alors le suivant (avant de centrer le résultat):
\begin{equation}\hat{f}_{X_S,ALE}(x_S)=\int\limits_{z_{0,1}}^{x_S}{\mathds{E}_{X_C|X_S}[\hat{f}^{(S)}(X_S,X_C)|X_S=z_S]dz_S}=\int\limits_{z_{0,1}}^{x_S}\int\limits_{x_C}\hat{f}^{(S)}(x_S,x_C)\mathds{P}(x_C|z_S)dx_Cdz_S\end{equation}
où $\hat{f}^{(S)}(x_S,x_C)=\frac{\partial \hat{f}(x_S,x_C)}{\partial x_S}$ est le gradient.

\paragraph{Estimation de l'ALE}
Décrivons à présent comment l'ALE est estimée en pratique, dans le cas où l'on veut comprendre le comportement d'une seule variable numérique $x_j$ ($S=\{j\}$) où $j\in \mathds{N}$.
 Nous divisons l'ensemble des valeurs prise par la variable $x_j$ en plusieurs intervalles, à savoir: $[z_{0,j},z_{1,j}]$, $[z_{1,j},z_{2,j}]$,...,$[z_{k_j(x)-1,j},z_{k_j(x),j}]$, avec $k_j(x)$ le nombre d'intervalles et  $z_{0,j}<z_{1,j}<...<z_{k_j(x),j}$.
Pour $k \in \{1,...,k_j(x)\}$,  notons $N_j(k)$ l'ensemble des individus de la base d'apprentissage pour lesquels la variable $x_j$ est dans l'intervalle numéro k: $[z_{k-1,j},z_{k,j}]$, et $n_j(k)$ le nombre d'individus dans $N_j(k)$.
Alors l'ALE au point $x\in \mathds{X}$ associée à la variable $j$ est estimée par la formule:
\begin{equation}
    \hat{\tilde{f}}_{j,ALE}(x)=\sum_{k=1}^{k_j(x)}\frac{1}{n_j(k)}\sum_{i:x_{j}^{(i)}\in{}N_j(k)}\left[f(z_{k,j},x^{(i)}_{\setminus{}j})-f(z_{k-1,j},x^{(i)}_{\setminus{}j})\right]
    \label{eq:ALE_1}
\end{equation}
Le terme d'effets locaux accumulés s'expliquent clairement sur cette formule : sur chaque intervalle, nous mesurons la différence de prédiction "locale", puis nous sommons sur tous les intervalles, afin d'avoir l'effet ``accumulé". En réalité, la véritable définition de l'ALE centre le terme précédent afin d'avoir un effet moyen à 0, il en découle la formule suivante :
\begin{align}
\hat{f}_{j,ALE}(x)&=\hat{\tilde{f}}_{j,ALE}(x)- \frac{1}{n}\sum\limits_{l=1}^{n}\hat{\tilde{f}}_{l,ALE}(x)\\ 
&=\sum_{k=1}^{k_j(x)}\frac{1}{n_j(k)}\sum_{i:x_{j}^{(i)}\in{}N_j(k)}\left[f(z_{k,j},x^{(i)}_{\setminus{}j})-f(z_{k-1,j},x^{(i)}_{\setminus{}j})\right] - \frac{1}{n}\sum\limits_{l=1}^{n}\hat{\tilde{f}}_{l,ALE}(x)
\label{eq:ALE_2}
\end{align}
Le fait de centrer la formule permet d'interpréter l'ALE comme l'effet d'une variable sur la prédiction en comparaison de la prédiction moyenne sur la base de données d'apprentissage.
Ainsi, si on obtient une ALE de $-2$ à un certain point $x$ lorsque $x_j=3$, cela signifie que lorsque la j-ième variable vaut 3,  la valeur de prédiction est inférieure de 2, en comparaison de la prédiction moyenne. 

Les intervalles dans la formule (\ref{eq:ALE_1}) sont généralement choisis comme étant différents quantiles de la distribution de la variable $x_j$ considérée. Cela permet d'avoir autant d'individus dans chaque intervalle, mais présente l'inconvénient d'avoir des intervalles de tailles très variables, notamment dans le cas où la queue de la distribution est lourde.
\section{Annexe : Alternatives de LIME}

\label{alternatives_LIME}
\subsection{Rappel sur LIME}
La procédure de LIME pour expliquer la prédiction faite par un modèle de type boîte noire, noté $b$, pour l'instance $x$ repose sur les étapes détaillées dans l'algorithme suivant :

\begin{itemize}
    \item La base d'apprentissage $X_{s_x}$ pour calibrer le modèle de substitution $s_x$ est construite en simulant des échantillons d'une loi normale pour chaque variable explicative $x_i$, avec la même moyenne et le même écart-type que les variables utilisées pour ajuster $b$.
   On note pour tout $j \in \{1,...,p\}, \ \mu_j=\frac{1}{n}\sum\limits_{i=1}^{n}x_{ij}$ (moyenne empirique de la variable $x_j$ dans la base d'apprentissage initiale) et $\sigma_j^2=\frac{1}{n-1}\sum\limits_{i=1}^{n}{(x_{ij}-\mu_j)^2}$ (variance empirique de la variable $x_j$ dans la base d'apprentissage initiale). On note aussi $n_{sim}$ le nombre d'individus utilisés pour ajuster $s_x$. Alors pour tout $j \in \{1,...,p\}$, on simule $n_{sim}$ échantillons indépendants de loi $N(\mu_j,\sigma_j^2)$. Ceci forme notre base $X_{s_x}$. Dans le cas de variables catégorielles, LIME travaille avec les probabilités de prédiction renvoyées par $b$.
   \item Le modèle substitué $s_x$ est ajusté en utilisant un modèle linéaire avec régularisation de type ridge, c'est-à-dire qu'on ajoute un paramètre de pénalité pour contrôler la variance du modèle. 
   Chaque instance $\tilde{x}$ de $X_{s_x}$ se voit attribuer un poids calculé par rapport à sa distance avec $x$, à l'aide d'une fonction noyau comme le noyau $RBF$ défini par $RBF(x,\tilde{x})=exp(-\frac{||x-\tilde{x}||^2}{2 \sigma^2})$, où $\sigma > 0$ est un paramètre à définir.
   Cette pondération a pour objectif de favoriser les points proches du point $x$ que l'on cherche à expliquer, afin d'avoir une interprétation locale.
   \item On génère des explications de la boîte noire $b$ en extrayant les coefficients de la régression linéaire mise en place avec le modèle $s_x$.
\end{itemize}

\subsection{LIVE}
L'article \cay{BreakDownLive} propose deux alternatives aux méthodes LIME et SHAP, appelées Live et BreakDown.
Nous nous concentrons dans cette partie sur la méthode Live (\emph{Local Interpretable Visual Explanations}).
L'objectif principal de cette méthode est le même que LIME, à savoir expliquer une prédiction particulière d'un modèle de type boîte-noire en utilisant un modèle de substitution local. La différence majeure avec LIME réside dans le choix des observations pour ajuster le modèle de substitution. L'algorithme de génération du voisinage est le suivant:
\begin{itemize}
    \item Input: $n_{sim}$, le nombre d'observations à générer et p,  le nombre de variables explicatives utilisées dans le modèle $b$ de boîte noire. Soit $x^* = (x^*_j)_{1\leq j\leq p}$ l'observation d'intérêt que l'on cherche à expliquer. 
    \item Initialisation :  dupliquer les observations données $n$ fois. Notons $X'$ la matrice contenant les données d'apprentissage du modèle de substitution : 
$X' = \underbrace{
\begin{pmatrix}
    x^*_1      & \cdots & x^*_1 \\ 
    \vdots &  & \vdots \\ 
    x^*_p      & \cdots & x^*_p 
\end{pmatrix}}_{n_{sim} \text{fois}}$.

    \item Pour $i \in \{1,...,n_{sim}\}$: Tirer $k \in \{1,...,p\}$ uniformément : $k\sim \mathscr{U}([\![1,n_{sim}]\!])$. Remplacer la $k$ i-ème variable avec un tirage de la distribution empirique (dans la base d'apprentissage) de cette variable: $X'[k,i] \sim \mathscr{L}(X_k)$.

\end{itemize}
La notation $\mathscr{L}(X_k)$ correspond à la loi "estimée" de la variable $X_k$ que l'on approche soit par une loi normale (paramètres trouvés par la méthode des moments par exemple), soit par tirage aléatoire de cette variable sur la base d'apprentissage, ce qui revient à réaliser des permutations sur la base initiale. Ces deux méthodes ont l'inconvénient de ne pas tenir compte de la dépendance entre les variables explicatives. Des alternatives existent mais ne sont pas abordées ici.
Toutes les observations (de $X'$) ainsi générées par l'algorithme ci-dessus sont supposées équidistantes à l'observation d'origine que l'on cherche à expliquer : aucune pondération n'est appliquée lors de l'ajustement du modèle de substitution local. Ceci permet de s'affranchir de la contrainte du choix de noyau de la méthode LIME.

\section{Annexe : autres graphiques utiles à l'analyse}
\label{annexe:graphes}

\begin{figure}[!h]
    \centering
    \includegraphics[scale = 0.9]{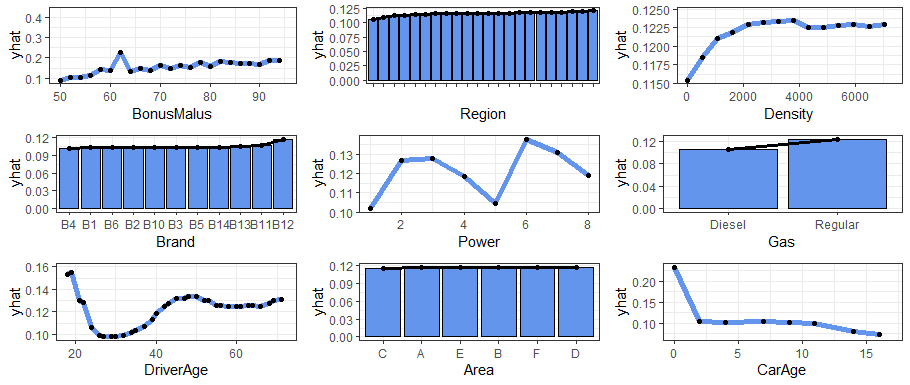}
    \caption{Graphiques de dépendance partielle ($PDP$) associés des différentes variables du modèle XGBoost $C$}
    \label{fig:pdp_all_xgb_freq_num}
\end{figure}

\begin{figure}[!h]
    \centering
    \includegraphics[scale = 0.9]{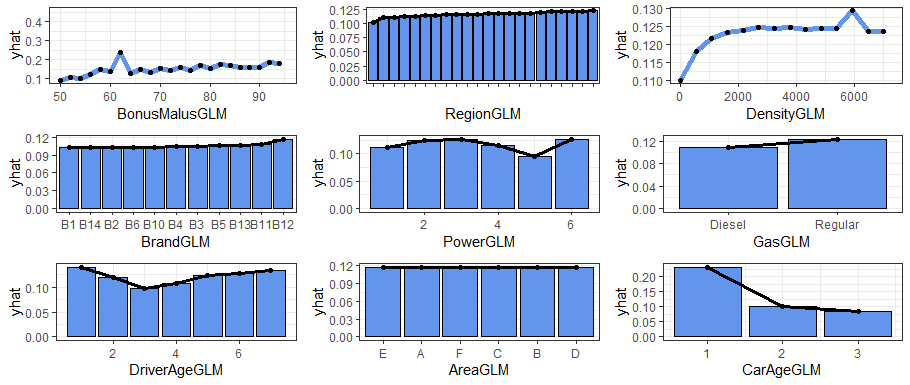}
    \caption{Graphiques de dépendance partielle ($PDP$) associés des différentes variables du modèle XGBoost $C-cat$}
    \label{fig:pdp_all_xgb_freq_cat}
\end{figure}

\end{document}